\documentclass[10pt,twocolumn,letterpaper]{article}

\usepackage{iccv}
\usepackage{times}
\usepackage{epsfig}
\usepackage{graphicx}
\usepackage{amsmath}
\usepackage{amssymb}
\usepackage{multirow}

\iccvfinalcopy 

\def\iccvPaperID{1929} 

\begin{document}

\title{Multiresolution hierarchy co-clustering for semantic segmentation\\ in sequences with small variations}
\author{David Varas \hspace{1.2cm} M\'onica Alfaro \hspace{1.2cm} Ferran Marques\\
Universitat Polit\`ecnica de Catalunya Barcelona Tech, Spain\\
{\tt\small \{david.varas,ferran.marques\}@upc.edu}
\thanks{This work has been done in the framework of the project BIGGRAPH-TEC2013-43935-R, financed 
by the Spanish Ministerio de Econom\'{i}a y Competitividad and the European Regional Development Fund (ERDF).}}

\maketitle


\begin{abstract}
   
This paper presents a co-clustering technique that, given a collection of images and their hierarchies, clusters nodes from these hierarchies to obtain 
a coherent multiresolution representation of the image collection. We formalize the co-clustering as a Quadratic Semi-Assignment Problem and solve it 
with a linear programming relaxation approach that makes effective use of information from hierarchies. Initially, we address the problem of generating 
an optimal, coherent partition per image and, afterwards, we extend this method to a multiresolution framework. Finally, we particularize this 
framework  to an iterative multiresolution video segmentation algorithm in sequences with small variations. We evaluate the algorithm on the Video Occlusion/Object Boundary Detection Dataset, showing that it produces state-of-the-art results in these scenarios.

\end{abstract}

\section{Introduction}
\label{sec:introduction}
%
%
%
The goal of co-clustering is to robustly segment a reference image (or various reference images) within a collection of closely related images 
(for instance, multiple views of a given scene or a video sequence with small variations) without any prior knowledge of the number of clusters.
This is closely related with the correlation clustering problem \cite{Bansal04}. 

Co-clustering approaches that model the problem as a Quadratic Semi-Assignment Problem \cite{Charikar2005} have been reported to 
outperform other co-clustering strategies \cite{Glasner2011}. However, such solutions present inconsistencies on the clusters propagation among 
images which prevent to obtain a coherent labeling through the collection of video images.

The goal of unsupervised video segmentation is to efficiently extract coherent groups of voxels from sequences to represent the video information with 
many less primitives.
Video segmentation techniques can be classified into three categories \cite{Corso2012}: (a) frame-by-frame processing, that leads to low temporal 
coherence results \cite{Brendel2009}; (b) iterative processing, that improves the temporal coherence while requiring reasonable algorithm complexity 
\cite{Paris2008}; and (c) 3D volume processing, that leads to the best results but implies high complexity algorithms and memory requirements 
\cite{Grundmann2010}.
\begin{figure}[t]
\centering
   \includegraphics[width=0.3\columnwidth] {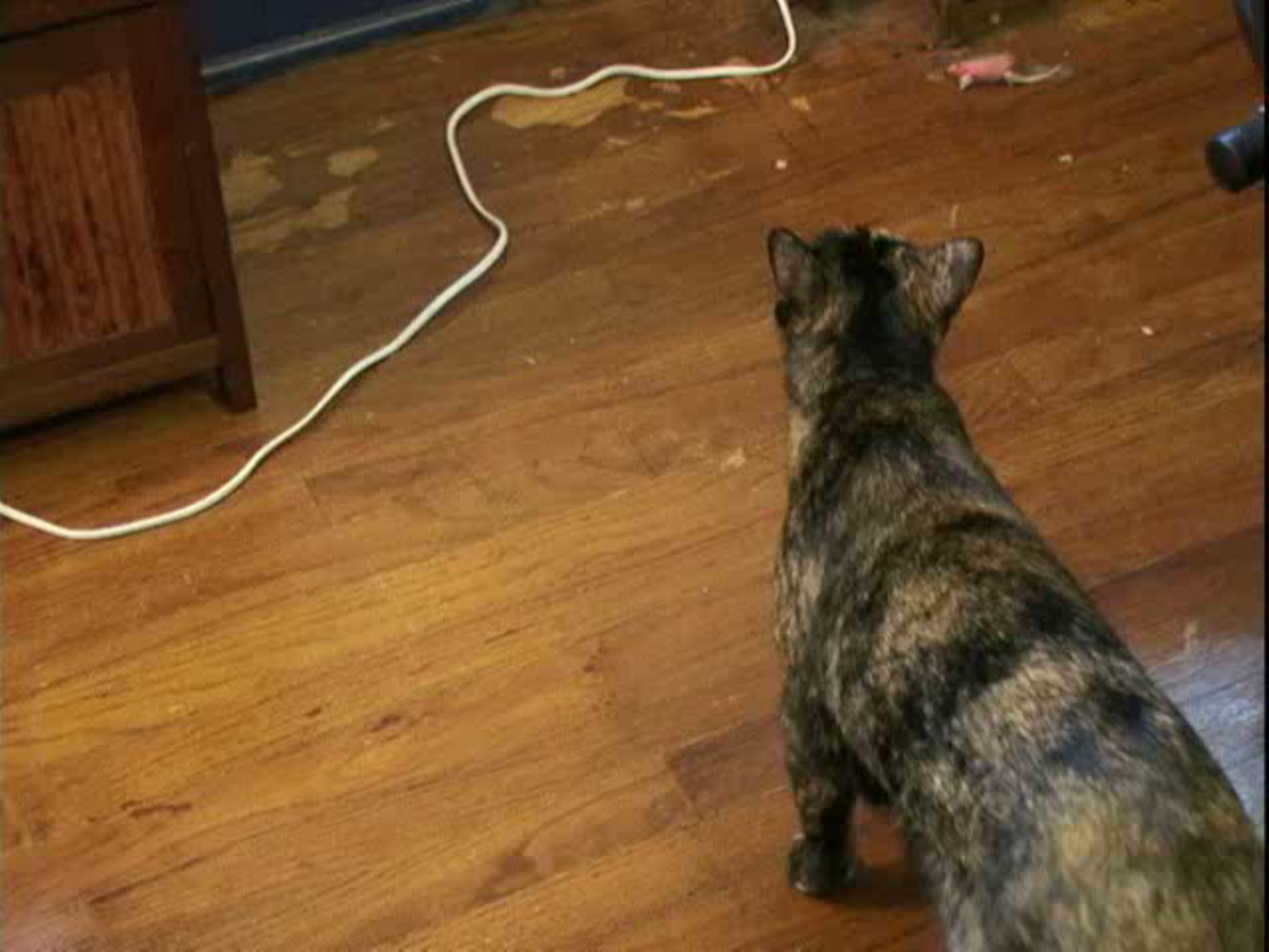}
   \includegraphics[width=0.3\columnwidth] {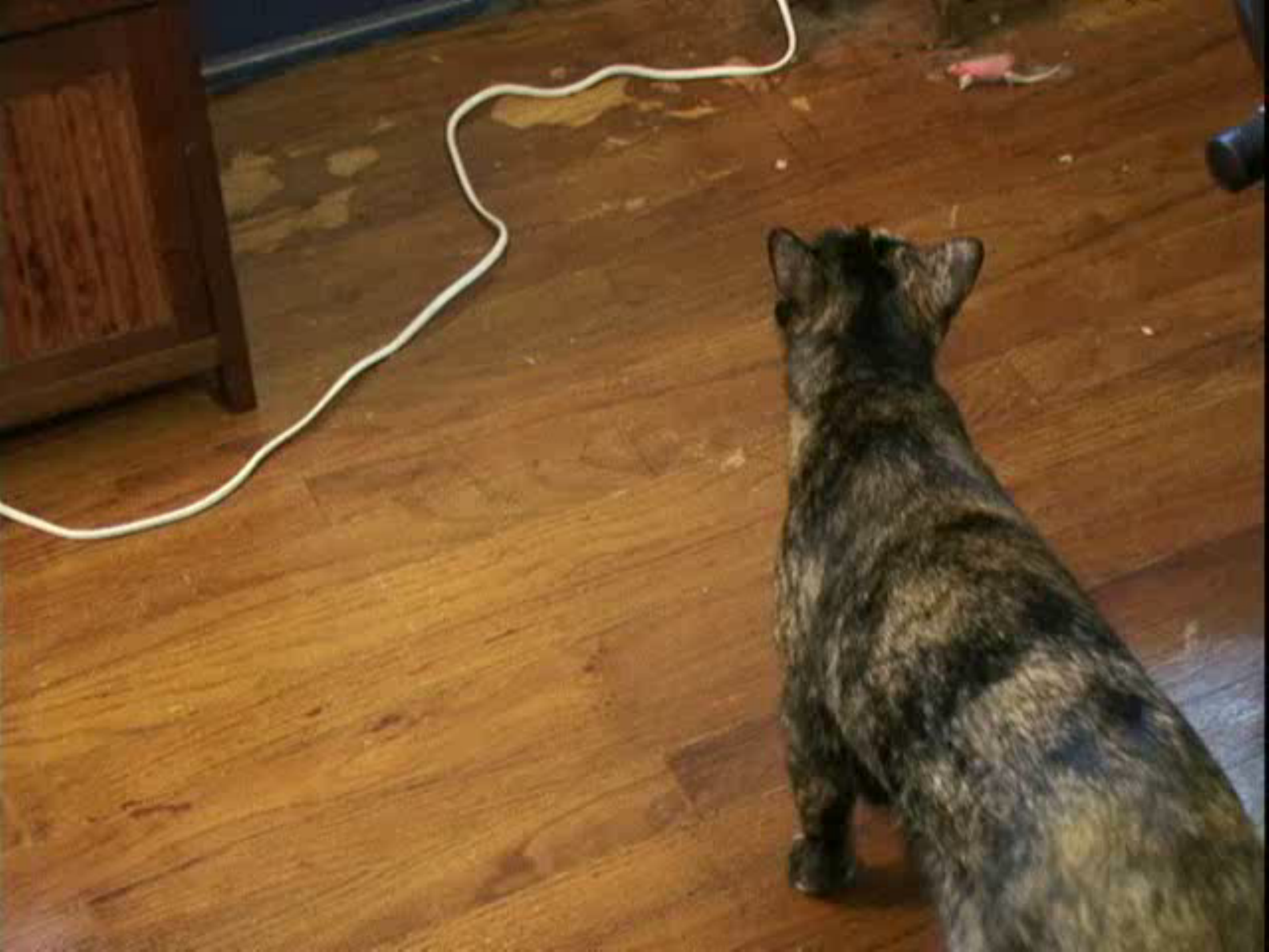}
   \includegraphics[width=0.3\columnwidth] {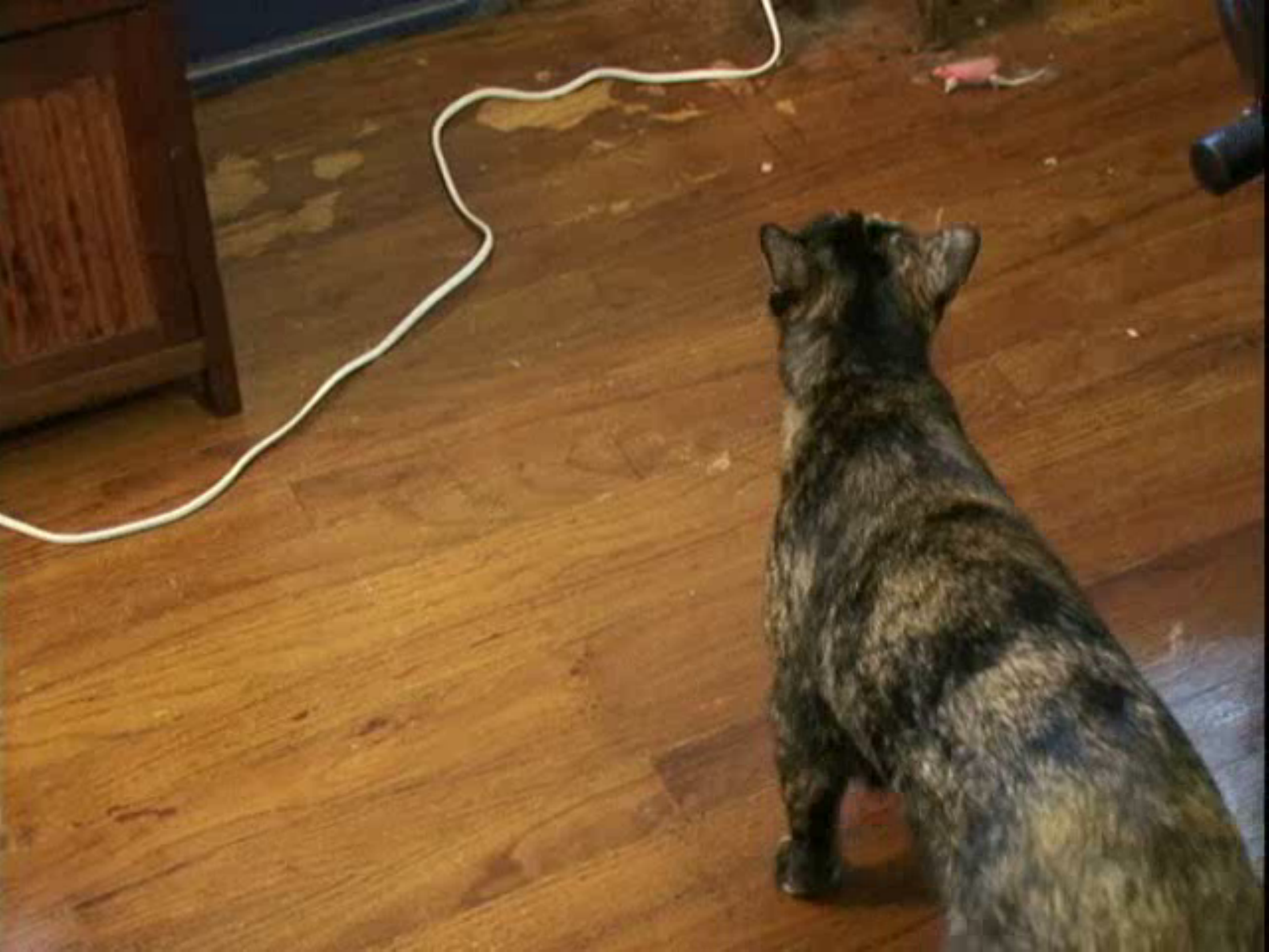}
\\
   \includegraphics[width=0.3\columnwidth] {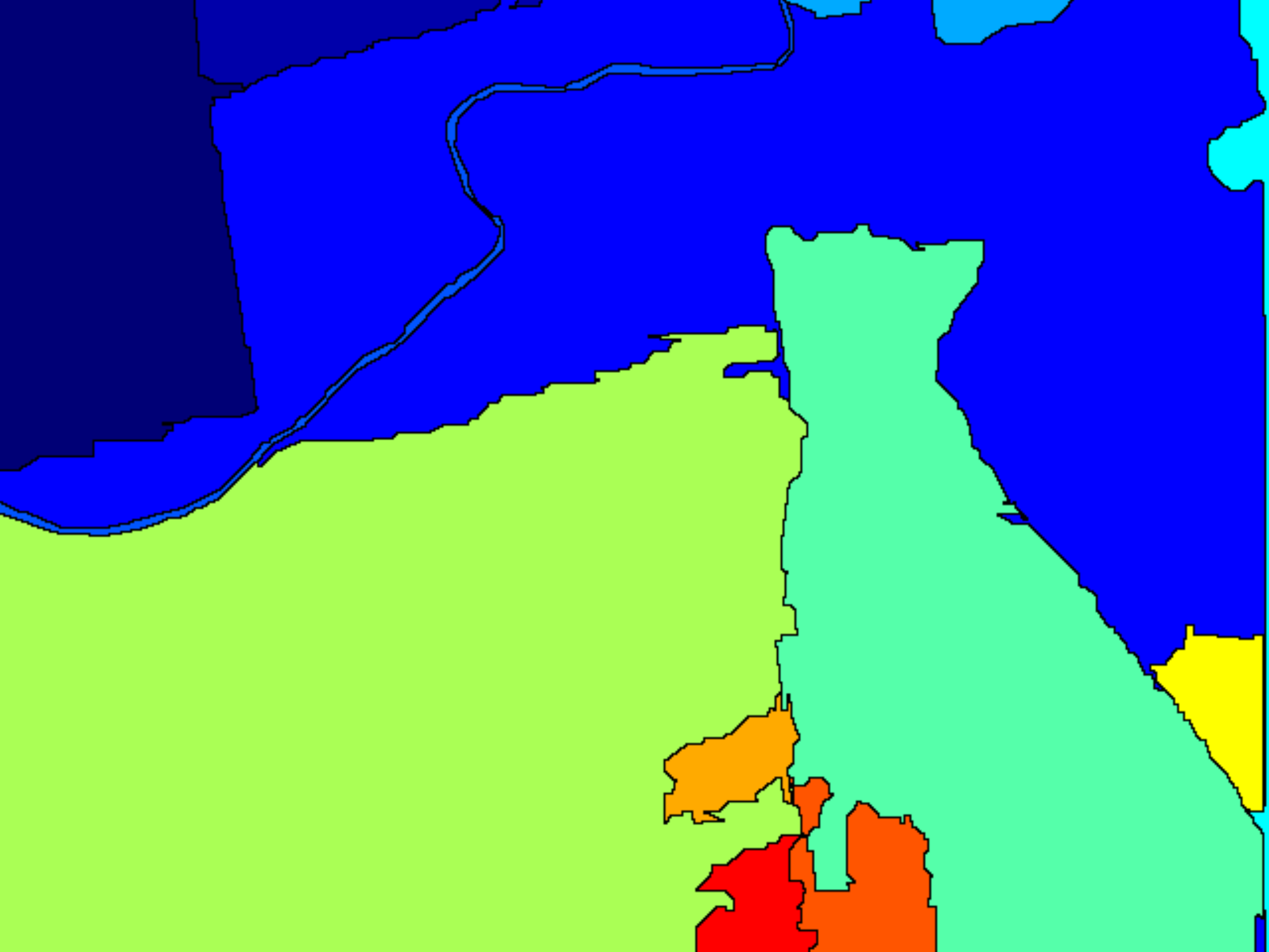}
   \includegraphics[width=0.3\columnwidth] {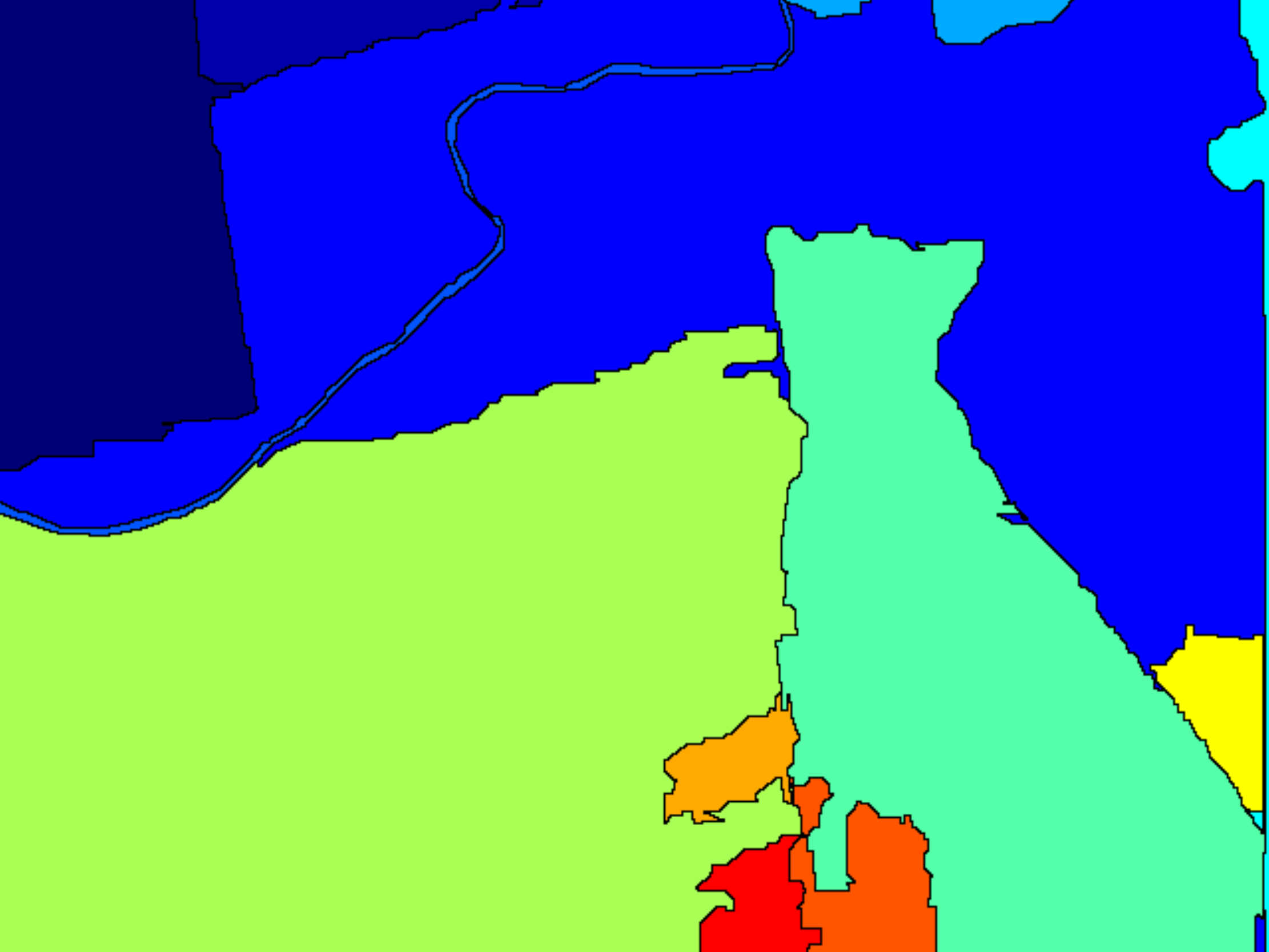}
   \includegraphics[width=0.3\columnwidth] {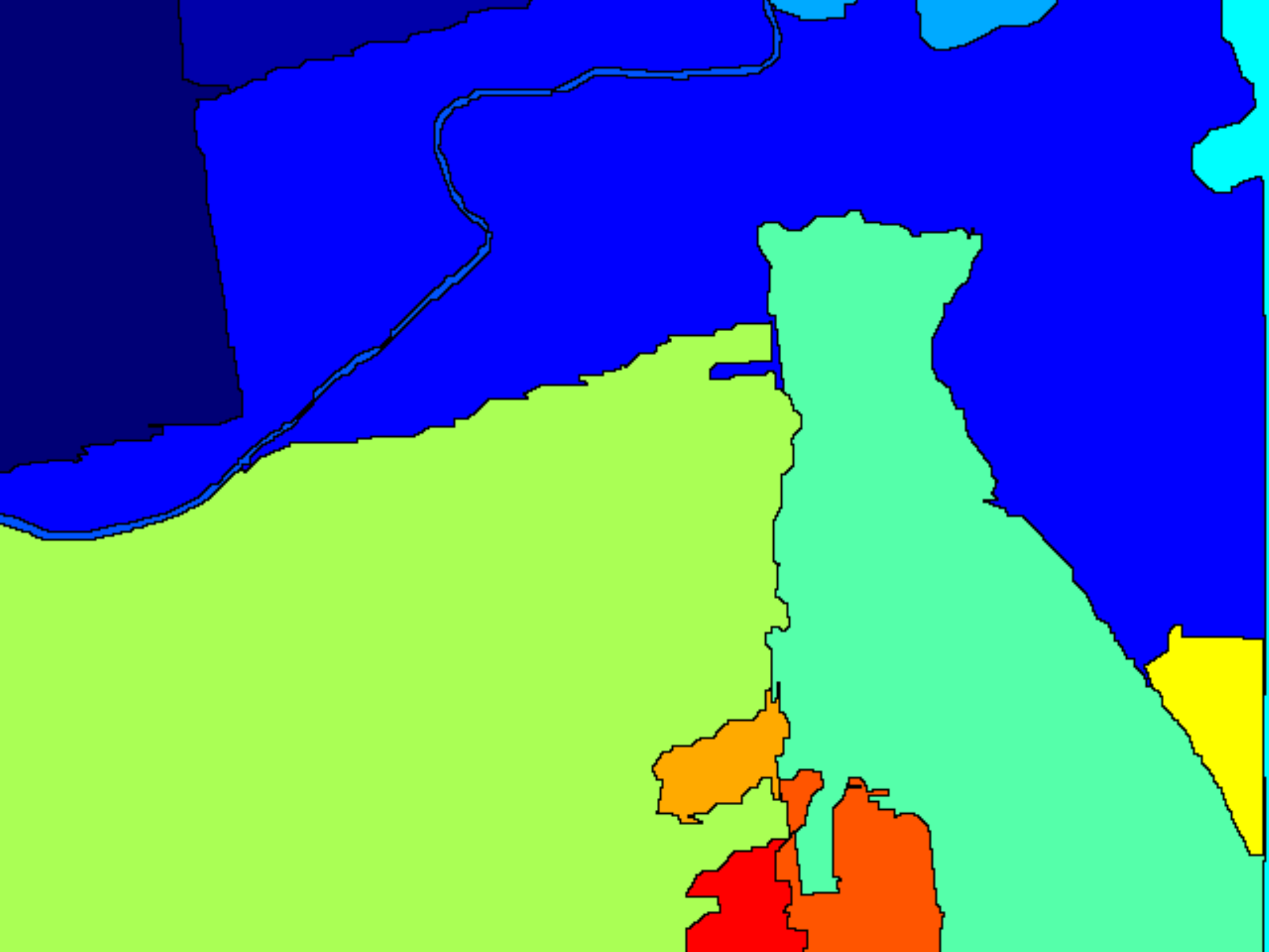}
\\
   \includegraphics[width=0.3\columnwidth] {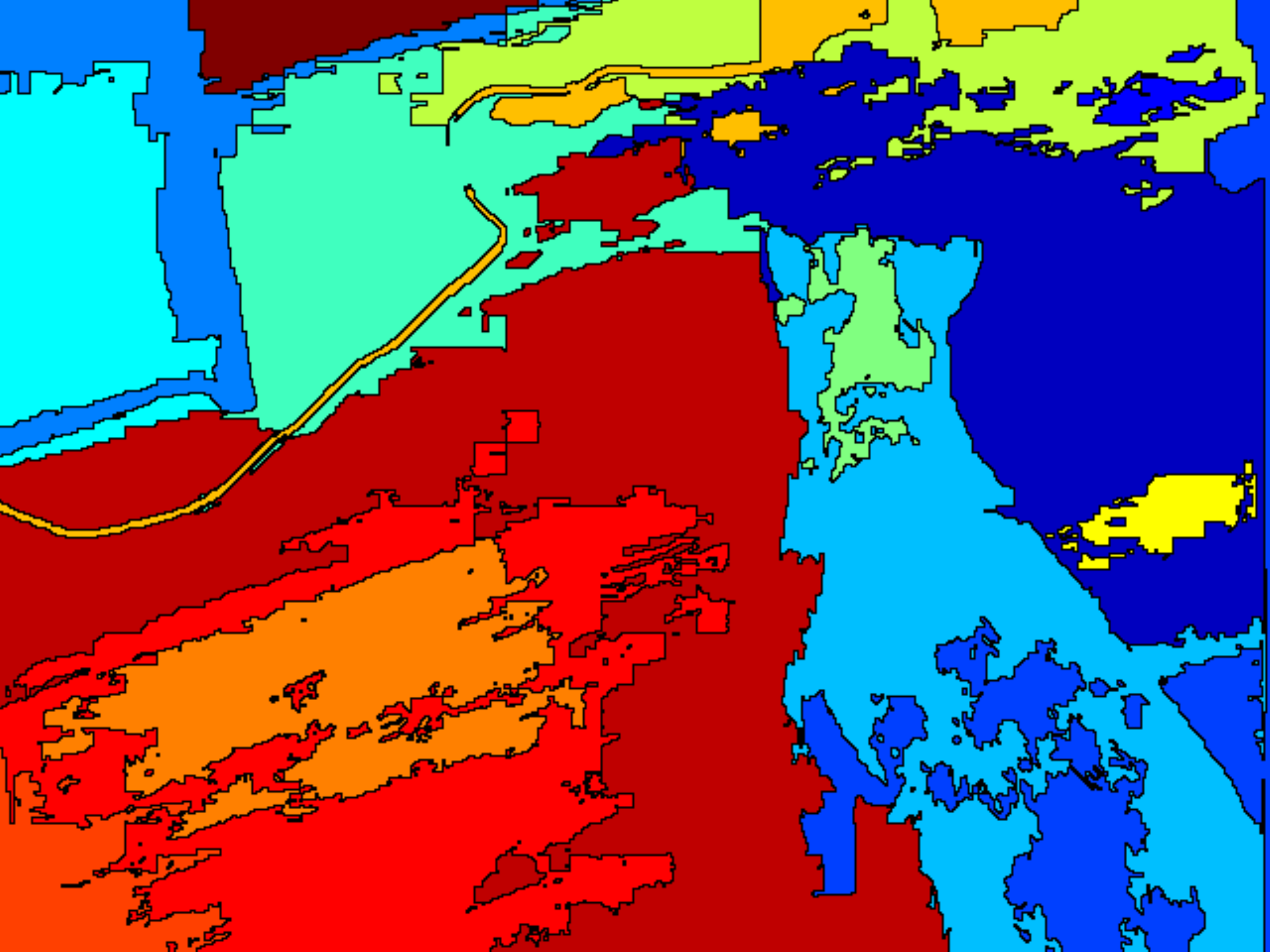}
   \includegraphics[width=0.3\columnwidth] {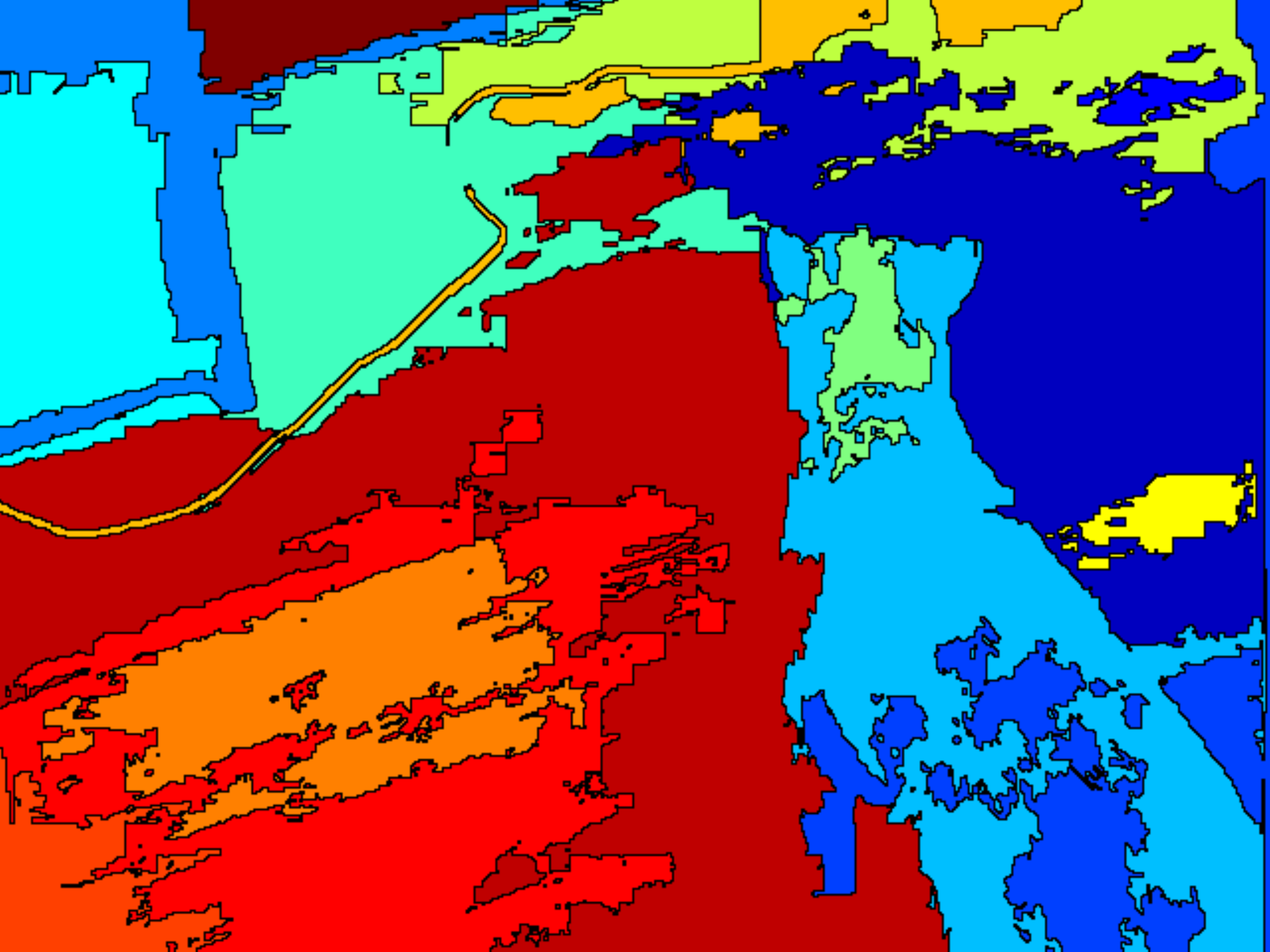}
   \includegraphics[width=0.3\columnwidth] {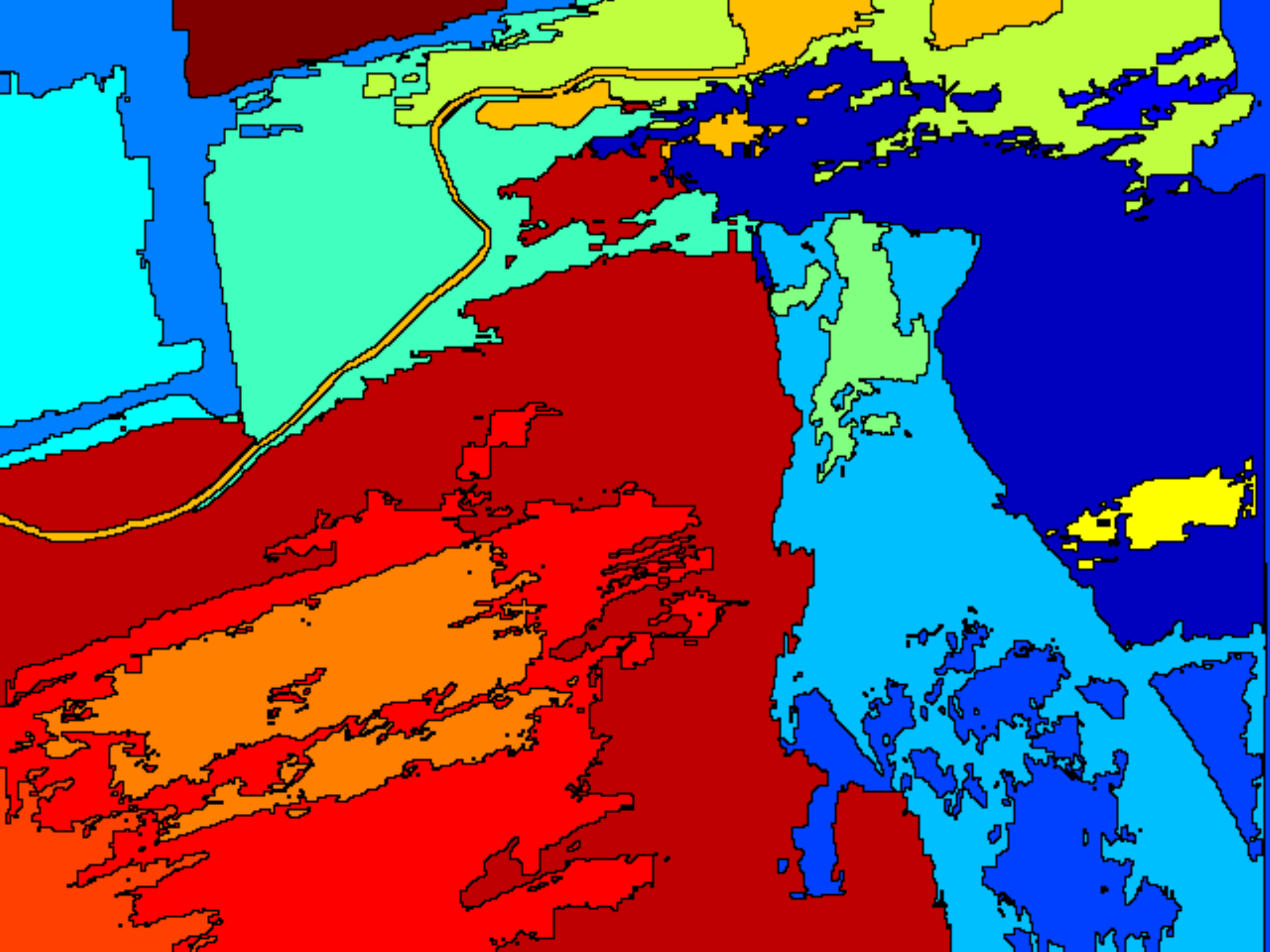}
\\
   \includegraphics[width=0.3\columnwidth] {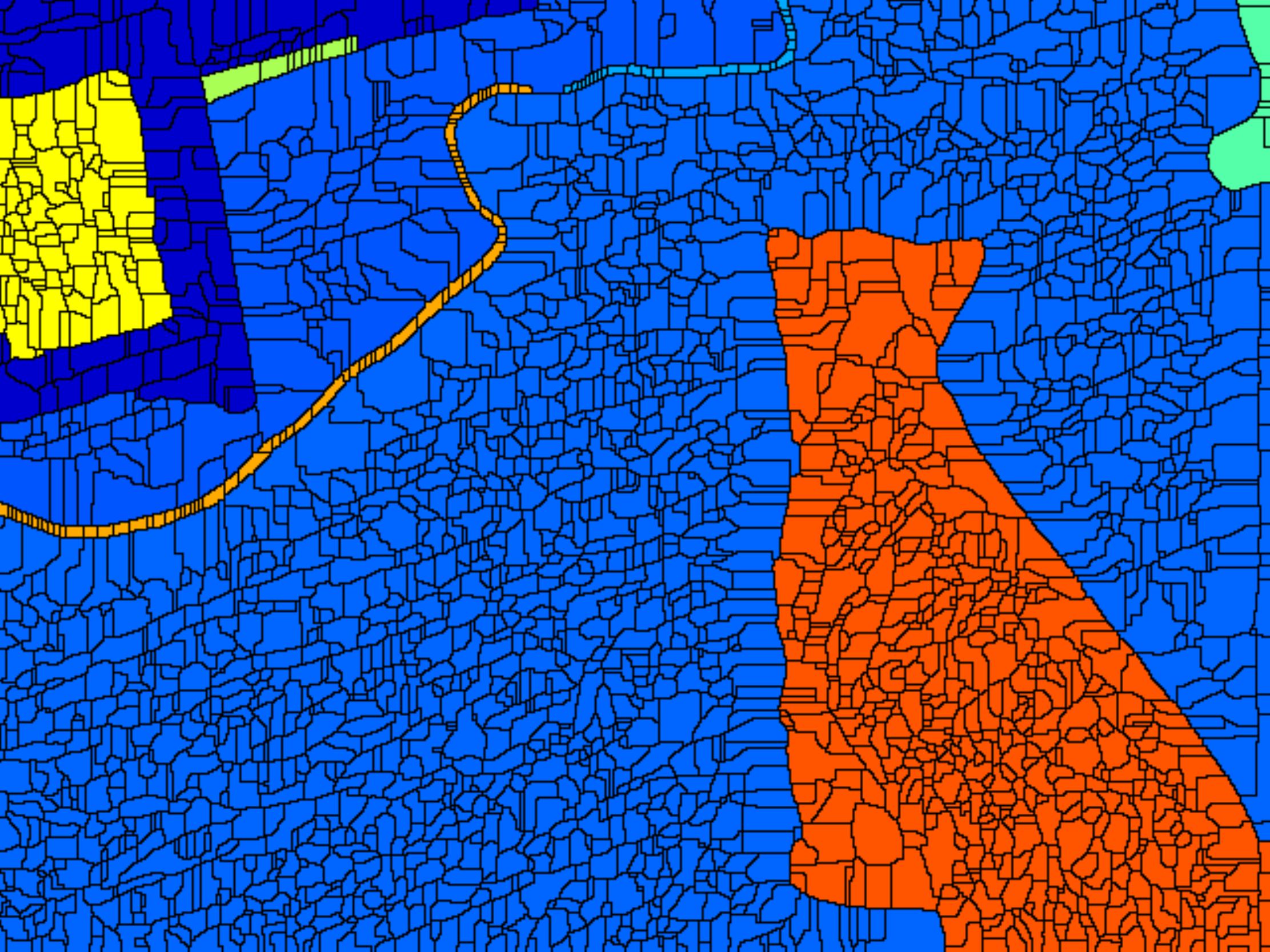}
   \includegraphics[width=0.3\columnwidth] {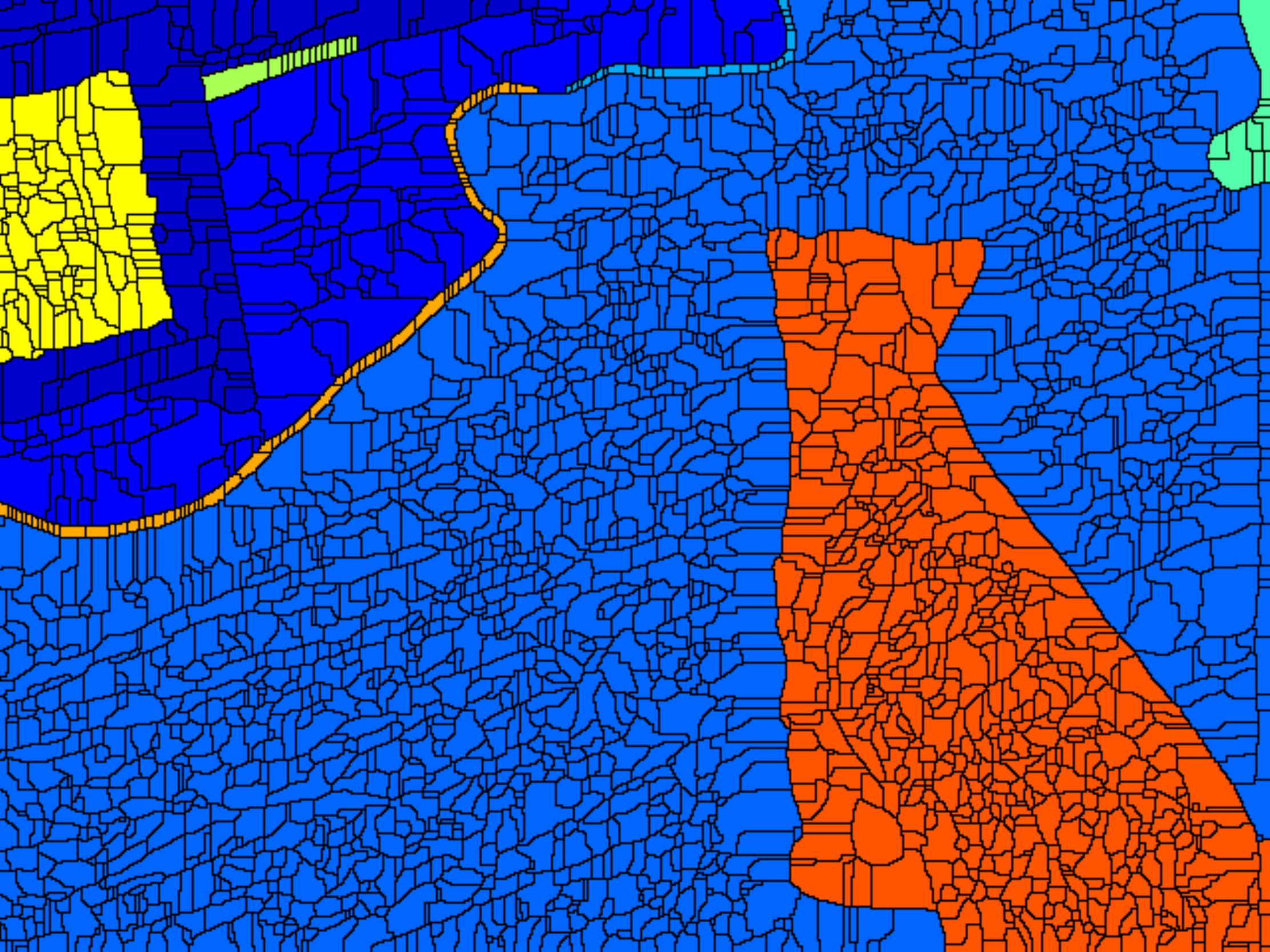}
   \includegraphics[width=0.3\columnwidth] {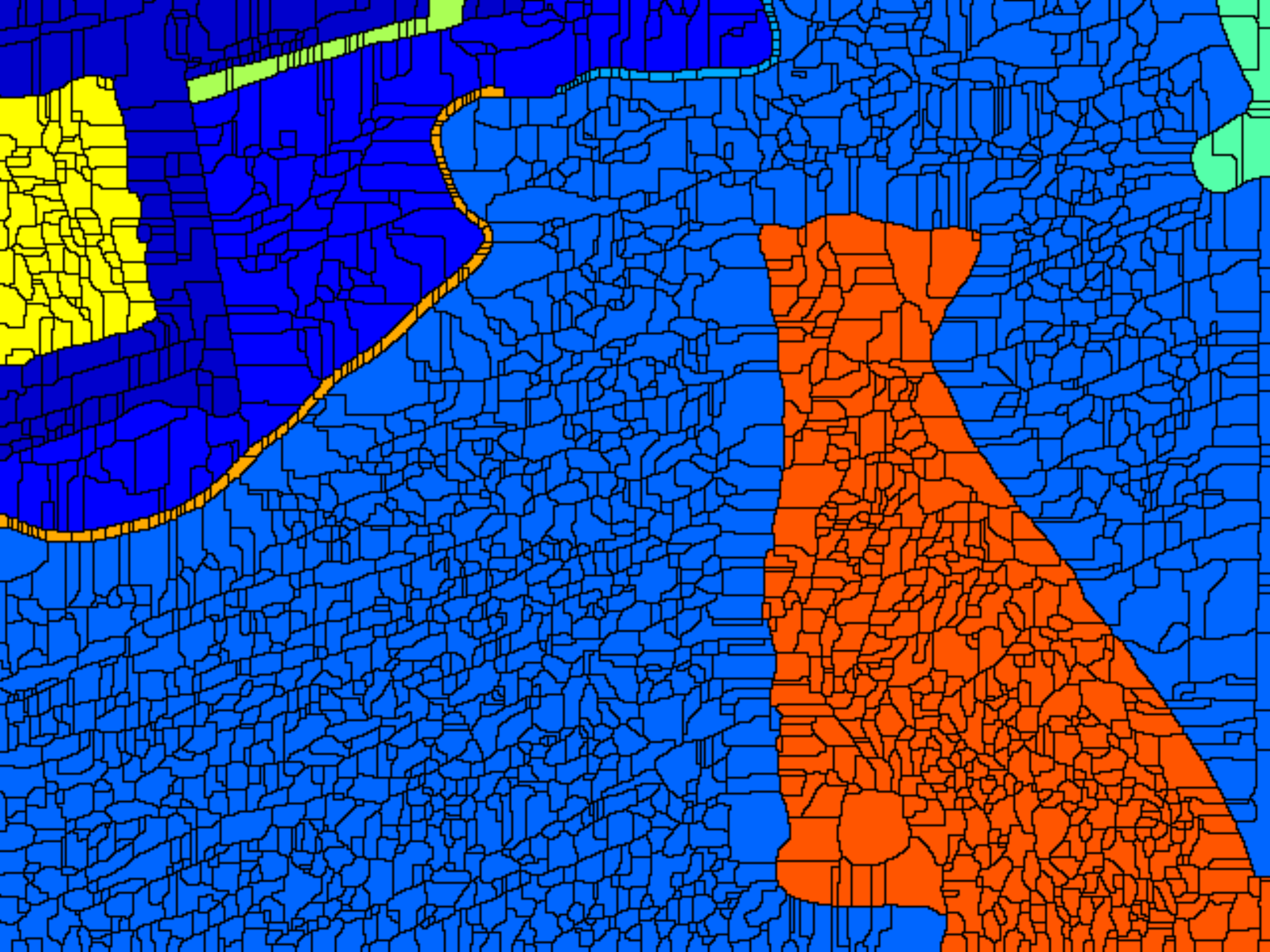}
\caption{Example of the video segmentation results on a sequence with little variation. Results are obtained with the minimum number of regions to 
achieve a given quality. First row: original frames of sequence \textit{zoe1} from 
the Video Occlusion/Object Boundary Detection Dataset. 
Second row: segmentation results of \cite{Grundmann2010}. Third row: segmentation results of \cite{Corso2012}. Fourth row: our results.}                                                                                         
\label{fig:imagesresults}
\vspace{-0.455cm}
\end{figure}
Regardless of the previous classification, it is nowadays widely accepted that multiresolution descriptions provide a richer framework for subsequent 
analysis, both in the image \cite{Malik2011} as in the video case \cite{Grundmann2010}. This way, current techniques mainly rely on motion information 
to build a set of coherent partition sequences, describing the video at different resolutions.
Video sequences with global motion or little variation in the scene pose problems to motion-based segmentation approaches. In these cases, to strongly 
rely on motion information does not help to infer the semantic in the scene. 
Figure \ref{fig:imagesresults} presents an example of this behavior.

To handle this kind of sequences, we propose a video segmentation method based on the co-clustering of a sequence of region-based hierarchical image 
representations. Moreover, we extend this co-clustering to produce a multiresolution representation of the video sequence. Our main contributions are:

\vspace{0.11cm}
\noindent An \textbf{optimization on hierarchies} that fully exploits the tree information avoiding inconsistencies of previous co-clustering approaches 
by coding the partitions with boundary variables and efficiently representing the hierarchical constraints (Section 4). 

\noindent An \textbf{iterative approach} for video segmentation based on the previous optimization process (Section 5), that combines the information at different resolutions. 

We conduct experiments on the Video Occlusion/Object Boundary Detection Dataset comparing with the techniques in 
(\cite{Grundmann2010}, \cite{Corso2012}, \cite{Galasso2012}, \cite{gunhee2012}, \cite{Joulin2012}). Comparisons are made using the implementations from respective 
author. We report an improvement in accuracy over state-of-the-art techniques. Figure \ref{fig:imagesresults}, fourth row shows an example of our results.

\section{Related work}
\label{sec:related_work}
In \cite{Grundmann2010}, a hierarchical graph-based method in which appearance and motion are used to group voxels is presented. This technique builds a coherent region-based representation of the entire video, processing it as a single stream. In our approach, we propose as well a multiresolution representation of the video sequence. Nevertheless, we avoid jointly processing the entire video and exploit the information provided by independent hierarchical segmentations. 

The concept of hierarchical graph-based video segmentation is also used in \cite{Corso2012}. In this work, sequences are processed relying on motion 
information and using bursts of frames in order to reduce the complexity of the algorithm. The information of these bursts is combined to create a 
supervoxel hierarchy of the entire video. Sequence partitions are then obtained using the uniform entropy slice in \cite{Corso2013}. In our work, we 
also process groups of images instead of the whole collection. Moreover, we iteratively propagate contour information at different resolutions.   

The work in \cite{Galasso2012} extends the hierarchical image segmentation of \cite{Malik2011} to the case of video, including motion information. To make the approach tractable, \cite{Galasso2014} proposes a spectral graph reduction which allows defining an iterative segmentation process for video streaming. In our work, although we present a global framework, we also propose an iterative segmentation process to make the problem tractable.

Previous techniques decrease their performance when scenarios with small variations are considered (Figure \ref{fig:imagesresults})
because motion does not help to describe semantics in the scene. To overcome this situation, we tackle the problem with a co-clustering approach. 
  



In the context of biomedical imaging, \cite{Vitaladevuni2010} stated a coclustering problem as a Quadratic Semi-Assignment Problem (QSAP) and, 
as in \cite{Charikar2005}, it tackled its solution with a Linear Programming (LP) relaxation approach. In \cite{Charikar2005}, the optimization 
function is computed from distances between regions and linear constraints are imposed on these distances. This relaxation creates a number of 
inequalities that grows as $O(n^3)$, where $n$ is the number of regions. 

In \cite{Vitaladevuni2010}, these constraints are only imposed over cliques in an adjacency graph on the regions. This approach bounds the number of constraints to $O(n^2)$. Moreover, a regularization parameter was introduced in \cite{Glasner2011} to avoid trivial solutions in the optimization process.
Although these approaches reduce the complexity of the problem, the solution of the optimization presents inconsistencies. These inconsistencies appear because the proposed constraints do not force the solution of the problem to be a partition. 

In our approach, we also define the co-clustering problem as a QSAP, but partitions are defined in terms of boundaries between regions. This allows us 
to reduce the complexity of the problem. Moreover, we substitute the previous constraints by imposing the structure of the hierarchies; this way, in addition to preventing inconsistencies, resulting partitions 
are closer to the semantic level.

Closely related to co-clustering between image partitions is the problem of co-segmentation, first introduced by \cite{Rother2006}. These methods take as 
input two or more images containing a common foreground object with varying backgrounds and attempt to segment the foreground object from the background. 
\cite{gunhee2012} extends the previous concept to the multiple foreground segmentation case. In it, the user has to define the number of 
background objects in the image collection and sets of adjacent regions (\textit{candidates}) are selected from an initial segmentation. To obtain a tractable problem, every set of regions is represented as a tree. In our case, we do not require any parameter and, for each image, a single hierarchy is computed.

Co-segmentation has also been applied to image sequences in a single resolution framework (\cite{Rubio2012}, \cite{Wei2013}) or using hierarchies \cite{Kim2012}. Note that co-segmentation algorithms would generally fail when tackling the case of scenes with small variations, since background in consecutive frames may also maintain its appearance. The work in \cite{Kim2012} proposes an optimization process over the nodes of the hierarchy. The use of nodes to define the inter image relations for all levels of the hierarchies would lead to an unfeasible number of variables and constraints. This problem is tackled in \cite{Kim2012} by restricting the inter relations to the highest level of the hierarchies. We solve that problem by defining the optimization process over boundary segments, 
which makes the problem tractable.

In this work, we propose a method to generate a multiresolution collection of coherent segmentations along a sequence with small variations. These 
segmentations are created clustering nodes from a set of non-coherent hierarchies associated with the video. This allows our technique to efficiently 
keep semantic contours at different resolutions and to eliminate random boundaries.    
\section{Working with hierarchies}

Each node of the hierarchy represents a region in the image, and the parent node of a set of regions represents their merging. For simplicity, let us 
assume that this hierarchy is binary (regions are merged by pairs). This structure is referred to as Binary Partition Tree in \cite{Salembier00}. 
Note that this assumption can be done without loss of generality, as any hierarchy can be transformed into a binary one. 

Commonly, such hierarchies are created using a greedy region merging algorithm that, starting from an initial \textit{leave partition} $P^1$, 
iteratively merges the most similar pair of neighboring regions. The concept of region similarity is what makes the difference among the various 
approaches.

\begin{figure}[h]
\centering
\includegraphics[width=1\columnwidth] {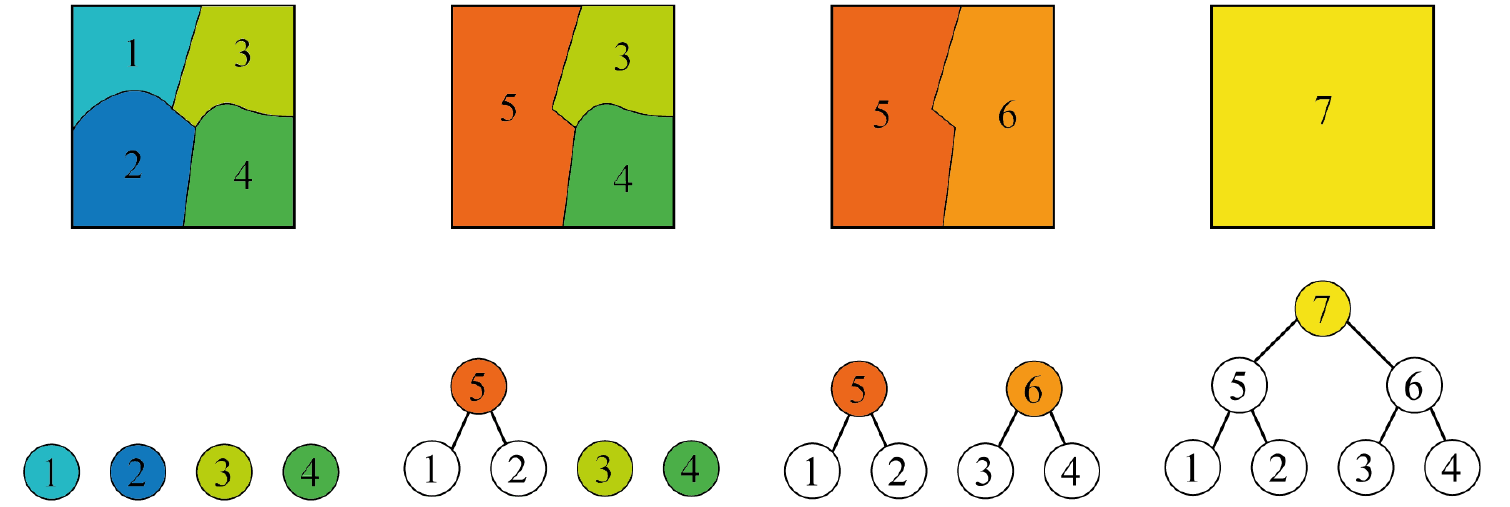}
\caption{Partitions generated with mergings of regions from the initial leaves partition $P^1$. The evolution of the hierarchy 
at each step is shown below the correspondant partition.}
\label{fig:bpt}
\end{figure}
The merging process ends when the whole image is represented by a single region, which is the root of the tree. The set of mergings that creates the tree, from the leaves to the root, is referred to as \textit{merging sequence}.

Given the previous example, let us define a vector $b=[ b_{1,2} \hspace{0.1cm} b_{1,3} \hspace{0.1cm} b_{2,3} \hspace{0.1cm} b_{2,4} \hspace{0.1cm} b_{3,4}]$ that encodes the boundaries between leaves. Using this notation, the partition generated after the first merging is represented by the sequence 
$[0 \hspace{0.1cm} 1 \hspace{0.1cm} 1 \hspace{0.1cm} 1 \hspace{0.1cm} 1]$, where $1$ represents an active boundary. 


In a binary hierarchy, a merging sequence contains $N^1$ partitions, where $N^1$ is the number of leaves (regions in $P^1$). This is the set of 
partitions that is usually analyzed when working with hierarchies. Still,  we generate partitions which may not be included 
in the merging sequence. For instance, in Figure \ref{fig:bpt}, the partition formed by $\{R_1, R_2, R_6\}$ would be generated and coded by the boundary combination 
$[1 \hspace{0.1cm} 1 \hspace{0.1cm} 1 \hspace{0.1cm} 1 \hspace{0.1cm} 0]$. This is done by analyzing all possible configurations of nodes in the hierarchy leading to a partition. Thus, we explore a larger number of contour combinations which allows us to use different resolutions at different parts of the image depending on its semantics.    
 
\vspace{-0.1cm}
\section{Co-clustering of hierarchies}
\label{sec:coclustering}
\begin{figure}[t]
\centering
\includegraphics[width=0.9\columnwidth] {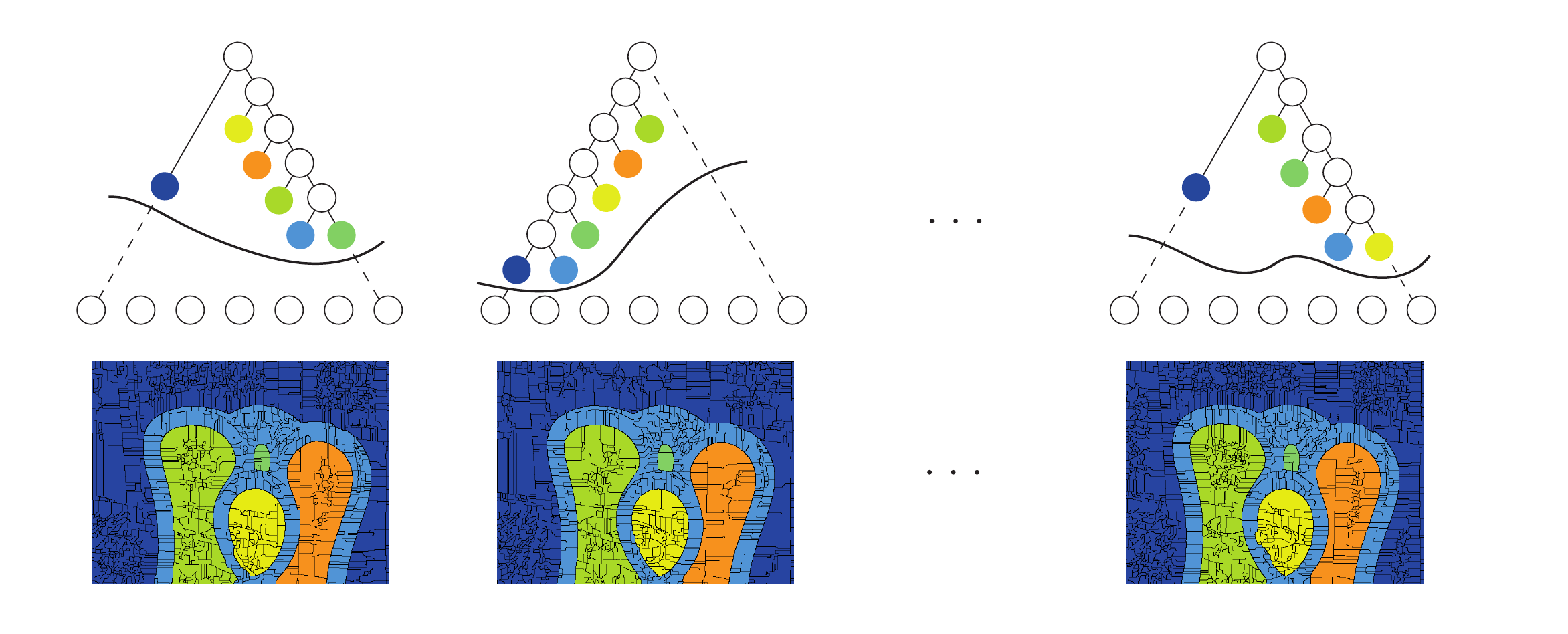}
\caption{Co-clustering of hierarchies from a collection of images. First row: nodes selected from the tree to create partitions. Second row: 
clusters created with unions of leaves describing tree nodes. Lines represent the cut in the tree producing the optimal partition. }
\label{fig:clustering}
\end{figure}

Let us assume that we have a collection of images, representing the same scene, which share a set of common contours but present a large number of 
random boundaries (e.g.: a video sequence with small variations or a multiple view scene representation). In this section we first present a global framework for, given such a collection of images and their associated 
and non-coherent hierarchies, obtaining a partition collection by clustering nodes from these hierarchies. This partition collection aims at keeping 
only the common contours and at producing coherent regions through the collection; that is, the various instances of the same object (or part) 
receive the same label in all the partitions of the collection (Figure \ref{fig:clustering}).

This is achieved by coding in the \textit{boundary matrix} the whole set of possible boundaries between adjacent regions in the collection. 
This matrix contains information about both the intra boundaries (between adjacent regions in the same image) and the inter boundaries (between 
adjacent regions in different images). The optimal boundary configuration (the co-clustering result) is achieved through an optimization problem 
that combines the boundary matrix information and the information about similarity between regions, which is coded in the \textit{similarity matrix}. 
As previously, the similarity matrix contains the information about intra and inter similarities between regions. Intra similarities are computed 
using global region descriptors while inter similarities rely on descriptors computed over all contour elements. To avoid inconsistencies in the result, some constraints are impossed to the optimization process. In our approach, intra constraints are obtained from the hierarchies, 
whereas the common triangular equations are adopted as inter constraints. In addition, we extend the previous hierarchical co-clustering to a multiresolution framework.

\subsection{Co-clustering problem definition}
\label{sec:problem_definition}
Formally, let us consider that we have a collection of M images $\{I_i\} = \{I_1, I_2, ..., I_M\}$ and their associated hierarchies 
$\{H_i\} = \{H_1, H_2, ..., H_M\}$. The merging sequence of a given hierarchy $H_i$ defines a set of partitions $\{P_i^p\} = \{P_i^1, P_i^2,...,P_i^{N_i^1} \}$, where $P_i^1$ is the \textit{leave partition} on which the hierarchy is built and $N_i^1$ is the number of regions in $P_i^1$. The $p$-th partition of hierarchy $H_i$ ($P_{i}^p$) is formed by a set of $N_{i}^p$ regions $ \{ R_{i}^{p,k} \}=\{R_{i}^{p,1}, ..., R_{i}^{p,N_{i}^p}\}$, where $\Psi \in \mathbb{R}^2$ and 
$\Psi = \cup_{k=1}^{N_{i}^p} R_i^{p,k}$ $\forall$ $p$.

To encode all possible partitions $\{\pi_i^q\}$ ( $\{P_i^j\} \subset \{\pi_i^q\}$) represented by a given hierarchy $H_i$, let us define its \textit{intra boundary} matrix, $B_{ii} \in {\{ 0, 1 \}}^{N_i^1 \times N_i^1}$. This is a binary matrix whose components are variables that relate all regions in $P_i^1$. This way,  $B_{ii}(m,n) = 1$ if, for the partition being coded, the boundary between leaves $m$ and $n$ is active; that is, if regions $m$ and $n$ have not been merged. 

Note that, by correctly zeroing some elements of this matrix, the whole set of partitions in $H_i$ ($\{\pi_i^q\}$) can be unequivocally described. This allows the co-clustering to fully exploit the richness of the hierarchical representation.

Boundaries between leaves of different partitions are coded in the \textit{inter boundary} matrices, 
$B_{ij} \in {\{ 0, 1 \}}^{N_i^1 \times N_j^1}$. Regions $m$ and $n$ from partitions $P_i^1$ and $P_j^1$ respectively belong to the same cluster if $B_{ij}(m,n)=0$.    

Then, a co-clustering between nodes from a collection of hierarchies is defined by a binary matrix, the \textit{boundary matrix}, $B$ $\in \{0,1\}^{N \times N}$ where 
$N = \sum_i N_i^1$. It encodes the intra and inter boundary information between leaves of the M images in the collection. 
\begin{align}
B=\begin{bmatrix}
       B_{11}      & ...     & B_{1M}           \\[0.3em]
       \vdots      & \ddots  & \vdots \\[0.3em]
       B_{M1}      & ...     & B_{MM}
\end{bmatrix}
\label{eq:boundaries_matrix}
\end{align} 

Note that $B$ only encodes the information of the leaves. The hierarchical information is introduced in the optimization process through 
the intra constraints (Section \ref{sec:intra_constraints}).

In practice, not all the variables represented in this matrix are usefull, as boundaries between non adjacent leave regions are not considered in the 
process. Thus, in contrast to previous partition-based approaches in which the number of constraints was bounded by $O(n^2)$ 
(\cite{Vitaladevuni2010}, \cite{Glasner2011}), our 
maximum number of intra constrains is proportional to $n$.

Our objective is to find the optimal boundary configuration that defines a collection of partitions $\{\pi_1^*, \pi_2^*, ..., \pi_M^*\}$ using nodes from hierarchies that are put in correspondace to form clusters. As proposed in \cite{Charikar2005}, the co-clustering can be stated as an optimization problem. To compact notation, let us define $B_{i,j}(m,n)=b_{m,n}$:
\begin{align}
 &\min_B \hspace{0.1cm} tr(QB) \hspace{0.3cm}\nonumber \\ 
 & s.t. \hspace{0.3cm} b_{m,n} \in \{0,1\}\hspace{0.2cm}\forall m,n \hspace{0.2cm} b_{m,m}=0
\label{eq:optimization_process}
\end{align} 
where $Q$ is a complex-valued Hermitian affinity matrix that measures the co-clustering quality. 

\subsection{Optimization Constraints}
As commented in Section \ref{sec:related_work}, we constrain  the optimization process using the information in the hierarchy to avoid the inconsistencies of previous approaches. Previous co-clustering techniques (\cite{Vitaladevuni2010}, \cite{Glasner2011}) use constraints that rely on the triangular equation to this purpose. This is, for each three-clique of adjacent regions, the labelling of these three regions to a single or to multiple clusters should be consistent. The main drawback of this approach is that label inconsistencies are only avoided in a reduced neighbourhood of each region. This information is expected to be propagated using the region adjacency, but inconsistencies are not specifically avoided out of this neighbourhood.    

In this work, as we perform co-clustering between hierarchies, we exploit the tree information to both encourage semantic fusions between 
regions and to reduce the number of constraints involved in the optimization.
\vspace{-0.2cm}
\subsubsection{Intra Constraints}
\label{sec:intra_constraints}
Each hierarchy $H_i$ contributes in two aspects to the optimization process. First, it defines the mergings between regions of its leave partition $P_i^1$ to form clusters. Second, it also includes the order in which these regions 
should be merged to represent each node of the tree. Note that this order is not conditioned by the merging sequence. These two contributions of the hierarchy information lead to a large number of constraints among the regions forming the subtree below a given node. Nevertheless, in this work, all these original constraints have been encoded with only two coupled constraints per node.

First, for a given parent node and in order to merge its two siblings, all the  leaves that form the boundaries between these two siblings 
should be merged. This is imposed by:  

\begin{equation}
\sum_{n\neq l}^{m,n} b_{m,n} = (N_c-1) b_{m,l}
\label{eq:intra1}
\end{equation}
where $N_c$ is the total number of common region boundaries from the leave partition that represents the union of both siblings, $m$ is a region from the first sibling and $n$, $l$ are regions from the second sibling. This condition imposes that all the variables representing boundaries between two siblings should have the same value.

Second, for a given parent node and in order to merge its two siblings, the  leaves that form their respective subtrees must also be merged:

\begin{equation}
\sum_{}^{n,l} b_{n,l} \leq N_m b_{m,o}
\label{eq:intra2}
\end{equation}
where $N_m$ is the total number of inner region boundaries from the leaves partition of both siblings, $m$ and $o$ are regions from the first sibling 
and $n$, $l$ are regions from the second sibling. This condition imposes that for a given node, a variable representing a boundary between two 
siblings can only impose a merging if all the leaves associated with the node are merged. 

Note that Equation \ref{eq:intra1} guarantes that all boundaries between two siblings are either active or non active at the same time. Therefore, the second constraint  \ref{eq:intra2}, coupled with the first one, ensures that the optimization process propagates the second condition to all the node boundaries.
\vspace{-0.1cm}
\subsubsection{Inter Constraints}
These constraints control the correspondances between nodes from different hierarchies. In this case, as we do not have any hierarchical relation 
for these nodes, the triangular equation is used to create the inter constraints: 
\begin{equation}
b_{m,n} \leq b_{m,l} + b_{l,n} \hspace{0.3cm} \forall e_{m,n}, e_{m,l}, e_{l,n} \in G
\end{equation} 
where $e_{m,n}$ is the edge between leaves $m$ and $n$ of the region adjacency graph $G$ computed from the leave partitions.

\subsection{Similarities}
Our co-clustering technique exploits the randomness of those partition contours that do not belong to semantic objects. In this process, the computation of region similarities is crucial to correctly match regions from different partitions. Two types of similarities are computed: \textit{intra similarities} (between leaves from the same hierarchy) and \textit{inter similarities} (between leaves from different hierarchies).

Previous clustering works in segmentation and cosegmentation frameworks (\cite{Glasner2011}, \cite{Kim2012}), use the color information to 
compute intra similarities. We propose to compute these similarities as:
\begin{equation}
W_{ii}(m,n) = \alpha_{m,n} \hspace{0.1cm} (1- e^{1-d_B(m,n)}) 
\end{equation}
where $\alpha_{m,n}$ is the length of the common boundary between leaves $m$, $n$ and $d_B(m,n)$ is the Bhathacharyya distance 
\cite{Bhattacharyya1943} of the 8-bin RGB color histograms of regions $m$, $n$.

Inter similarities are used to create clusters combining nodes from different hierarchies. In \cite{Glasner2011}, inter similarities are computed using a HOG-based descriptor. Although this gradient information may be enough in some cases, additional 
descriptors able to robustly match region contours are required. However, only those descriptors that can be efficiently computed should be taken into account.


We propose to combine three simple yet effective descriptors, which are computed over the contour elements 
of each partition. These descriptors are combined in a feature vector associated with each contour element, what allows us to keep the 
additivity property that is the key to formulate our problem as a linear optimization.

Inter image similarity between regions $m$ and $n$ from partitions $P_{i}^1$ and $P_{j}^1$ respectively should be 
proportional to their joint probability $p(m,n)$. We considere three types of information to model differences between 
regions from different partitions: changes of \textit{color/illumination}, \textit{deformations} and small changes of \textit{position}. 
In terms of probability, we consider these processes to be independent:
\begin{equation}
 p(m,n) =  p_C(m,n)  p_D(m,n) p_P(m,n)
\end{equation}

The \textit{color information} is obtained from a histogram of pixels in a neighborhood of the boundary elements. 
Two histograms are computed in the direction of the normal to the contour element (one in the analyzed region and the other in the adjacent region) 
and they are averaged.
To handle possible \textit{deformations}, shape information around each contour element is captured with a HOG descriptor. 
In our work, HOGs are computed using the gPb \cite{Maire2008} information. Finally, 
\textit{position} changes are captured with the Euclidean distance between elements.

Similarity between contour elements is computed as 
$W_{ij}(u,v) = e^{(f_i^u-f_j^v)^T \Sigma^{-1} (f_i^u-f_j^v)}$, where $f_i^u$ is the feature vector of contour element $u$ that belongs to $P_i^1$. This vector is formed as the concatenation of the three types of descriptors previously described. We allow matchings between contour elements that are closer than 20 pixels. Otherwise, $W_{ij}(u,v)=0$.

Once both inter and intra similarities are computed for all contour elements of the leave partitions, a similarity matrix between regions is built for each pair of hierarchies. 
\begin{equation}
 Q_{ij} = {O_i}^H W_{ij} {O_j}
\end{equation}
where $O_i$, $O_j$ are complex matrices that describe the edges orientations (computed using the gPb \cite{Maire2008} information) of all  contour elements from partitions $P_i^1$ and $P_j^1$,
and $W_{ij}$ encodes the inter similarities between these elements.

Finally, the similarity matrix $Q$ that measures the quality of the co-clustering is built using the information of all the inter and intra 
similarity matrices as in Equation \ref{eq:boundaries_matrix}.

\subsection{Optimization process}
Using the similarity matrix and the constraints presented in this section, the optimization process of Equation \ref{eq:optimization_process} 
can be formulated as:
\begin{align}
\min_B &\sum_{m,n} q_{m,n}b_{m,n} \nonumber \\ 
 \hspace{0.2cm} s.t. \hspace{0.4cm} &b_{m,n} \in \{0,1\} \hspace{0.4cm} b_{m,n} = b_{n,m} \hspace{0.4cm} \forall n,m \hspace{0.2cm}\nonumber \\
                      \sum_{n\neq l}^{m,n} b_{m,n}& = (N_{c}-1) b_{m,l} \hspace{0.1cm},\hspace{0.1cm} \sum_{}^{n,l} b_{n,l} \leq N_{m} b_{m,o} \hspace{0.1cm} \forall \mathfrak{p} \in \{H_i\} \nonumber \\ 
                     b_{m,n} &\le b_{m,l} + b_{l,n} \hspace{0.6cm} \forall e_{m,n},e_{m,l},e_{l,n} \hspace{0.05cm} \in \hspace{0.05cm} G 
\label{eq:optimization_process_with_constraints}
\end{align}
where $\mathfrak{p}$ represents any parent node in the collection of hierarchies. The result of this optimization is a binary matrix $B^*$ that describes the 
collection of optimal partitions $\{\pi_1^*, \pi_2^*, ..., \pi_M^*\}$. Thus, nodes from the collection of hierarchies $\{H_i\}$ have been clustered 
with the same label and semantic contours are preserved through the collection. 

\subsection{Multi-resolution}

\begin{figure}[t]
\centering
\includegraphics[width=0.9\columnwidth] {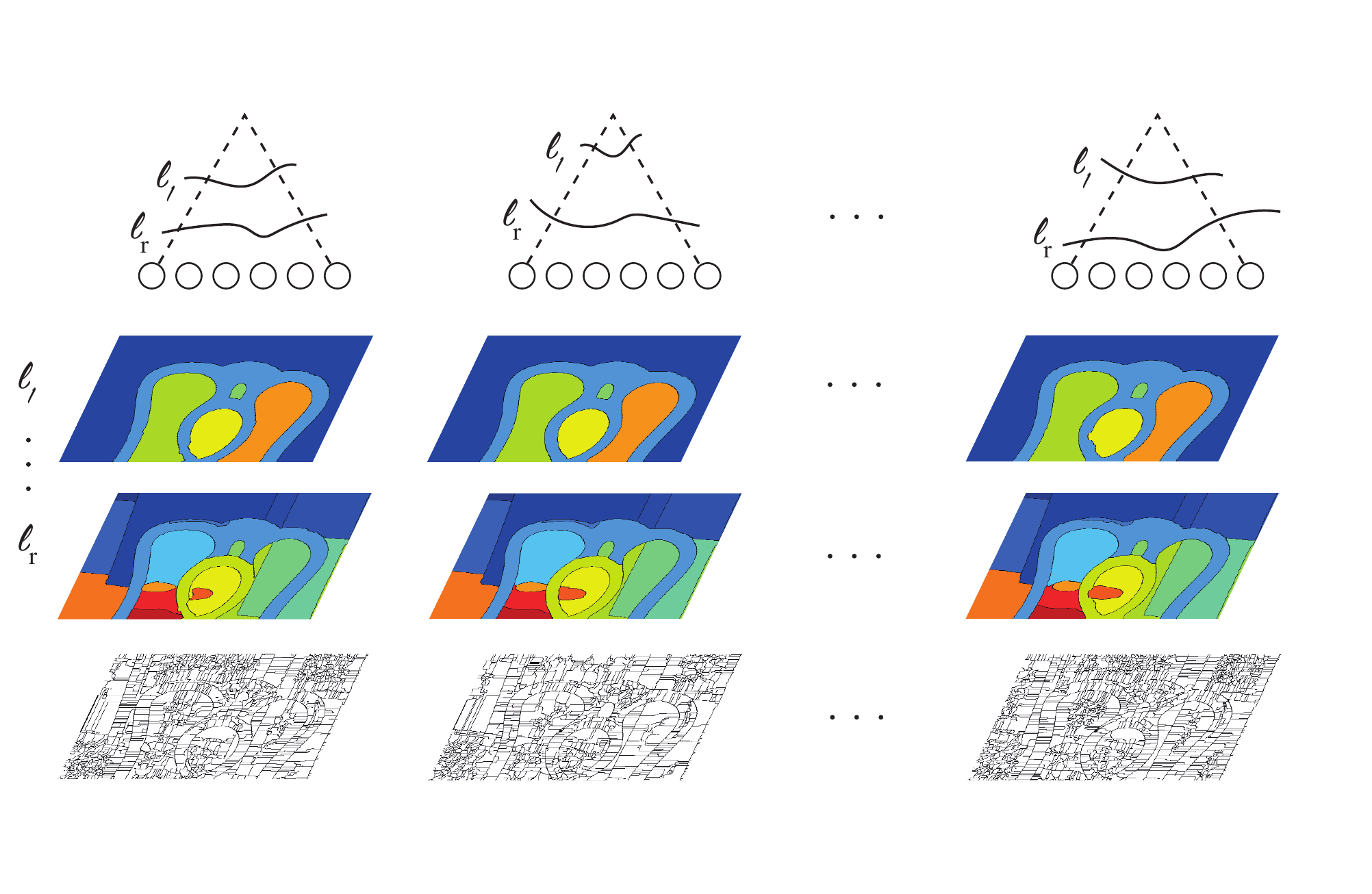}
\caption{Multiresolution hierarchy co-clustering of an image collection. First row: different cuts of each tree associated with different 
resolutions. Second and third row: optimal partitions generated by the previous hierarchy cuts. Fourth row: leave partitions }
\label{fig:multiresolution}
\end{figure}

Nowadays, it is commonly accepted that multiresolution region-based descriptions provide a rich framework for image and video analysis 
\cite{Arbelaez2014}, \cite{Grundmann2010}. In this section, we extend the previous hierarchical co-clustering to a multiresolution framework as it 
is illustrated in Figure \ref{fig:multiresolution}. 

This is, for each hierarchy involved in the optimization process ($H_i$), we cluster nodes to obtain $N_r$ partitions, forming a new optimal hierarchy ($\mathcal{H}_i^*$) that represents the image at $N_r$ different resolution levels ($\mathcal{H}_i^* = \{\pi_i^{1*}, \pi_i^{2*}, ..., \pi_i^{N_r*}\}$). 
Moreover, the collection of optimal partitions generated for each resolution should keep their inter correspondances. 


Let us consider a clustering problem as presented in Equation \ref{eq:optimization_process_with_constraints}, from which a boundary matrix $B$ is obtained for each generated partition. The number of active boundaries in $B$ has a direct relation with the resolution of the resulting partitions and, in particular, that of intra boundaries. When imposing in the optimization process a low (high) number of intra contours, coarser (finer) resolutions are obtained. We have observed that parameterizing the search in the solution space with respect to the number of intra contours allows the algorithm to produce a set of well distributed resolutions.
Formally, given a collection of hierarchies ($\{H_i\}$), their nodes are clustered to form a collection of partitions of a given resolution ($ \{\pi_1^{r*}, \pi_2^{r*}, ..., \pi_M^{r*}\} $) by constraining the optimization problem presented in Equation \ref{eq:optimization_process_with_constraints} with an additional condition for each hierarchy: 

\begin{equation}
(T_{r}- \beta) \cdot N_{b} \leq \sum_{}^{m,n} b_{m,n} \leq T_{r} \cdot N_{b}
\label{eq:multiresolution_constraint}
\end{equation}
where $N_b$ is the number of active boundaries to encode the leave contours, $T_{r}$ is the maximum fraction of these contours to describe the 
$r$-th coarse level and $\beta$ represents the maximum difference in number of boundaries between consecutive levels.    

This approach allows two search strategies. When $\beta = T_{r} - T_{r-1}$, a complete set of consecutive, equal sized subspaces is analyzed. On the contrary, 
when $\beta \hspace{0.1cm} \textless \hspace{0.1cm} T_{r} - T_{r-1}$ a coarser sampling of the solution space is performed.

\section{Multi-resolution video co-clustering}
\label{sec:video_co_clustering}
\begin{figure}[t]
\centering
\includegraphics[width=\columnwidth] {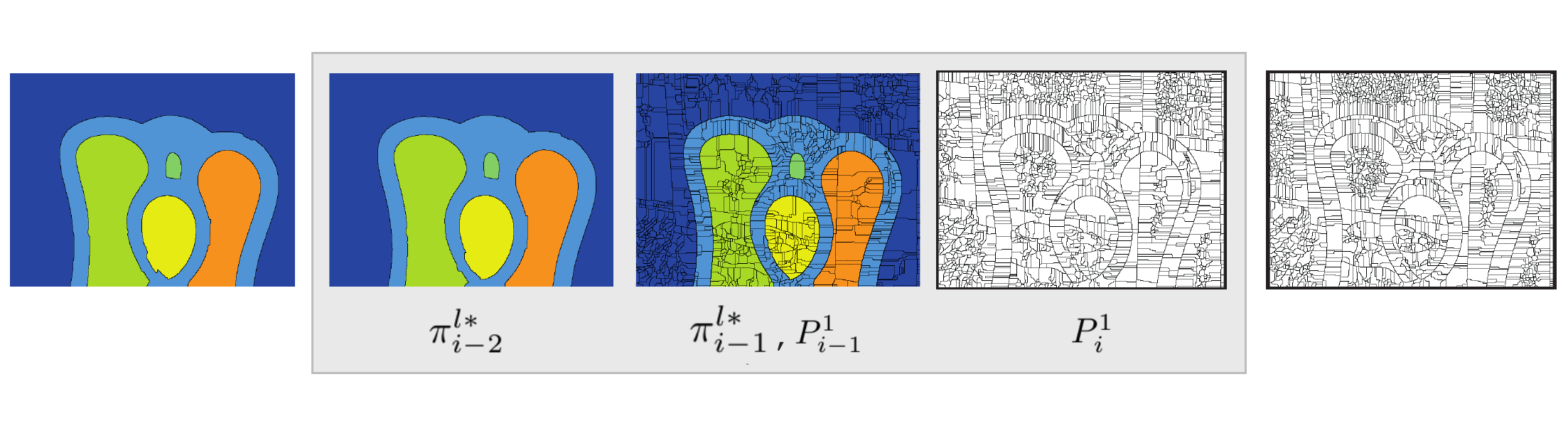}
\caption{Iterative algorithm to propagate semantic information through a video. As it can be seen, information of different coarse 
levels ($\pi_{i-2}^{l*}$, $\pi_{i-1}^{l*}$, $P_{i-1}^1$) is used to compute the optimal current frame partition without modifying the previous 
results. }
\label{fig:sequence}
\end{figure}

In this section we propose to particularize the technique presented in Section \ref{sec:coclustering} to a multiresolution video 
segmentation algorithm for sequences with small variations. Note that the previous co-clustering technique could be adapted to a 3D volume approach, 
as in \cite{Grundmann2010}. However, such an approach would require high memory resources (Section \ref{sec:introduction}). 
Thus, we adopt an iterative approach as in \cite{Galasso2014} (Figure \ref{fig:sequence}).


We propose to propagate clusters along sequences at various resolutions, taking into account the information in previous processed frames. As in \cite{Corso2012}, we use pieces of video and propagate the result through the sequence. In our case, we propagate semantic contours using information from different granularities in the optimization process.  This is a forward-only online processing, and the results are good and efficient in terms of 
time and complexity.

In particular, for each image ($I_i$) in the sequence and for a given resolution ($r$), we perform a joint hierarchical co-clustering with the 
clustering result of the two previous frames at two different scales: the resolution level under analysis and the leave partition scale (see Figure \ref{fig:sequence}). Precisely, we construct the boundary matrix $B$ using the optimal partition in $i-2$ at level $r$ ($\pi_{i-2}^{r*}$) and the leave partitions in $i-1$ and $i$ ($P_{i-1}^1$ and $P_{i}^1$). 

Moreover, the optimization problem in \ref{eq:optimization_process_with_constraints} and  \ref{eq:multiresolution_constraint} is further constrained 
imposing two additional conditions. In order not to modify previous co-clustering results, regions in $\pi_{i-2}^{r*}$ must not be merged
\begin{equation}
\sum_{}^{m,n} b_{m,n} = N_v
\end{equation}
where $b_{m,n}$ are intra or inter boundary variables from $\pi_{i-2}^{r*}$ and $P_{i-1}^1$ that encode the boundaries between clusters 
of $\pi_{i-2}^{r*}$ and $\pi_{i-1}^{r*}$, and $N_v$ is the cardinality of these variables.

In turn, regions in $P_{i-1}^1$ must be merged to form $\pi_{i-1}^{r*}$ and inter correspondances between clusters must be kept:
\begin{equation}
\sum_{}^{m,n} b_{m,n} = 0
\end{equation}
where $b_{m,n}$ are intra or inter boundary variables from $\pi_{i-2}^{r*}$ and $P_{i-1}^1$ that encode the unions of inter and intra clusters of 
$\pi_{i-2}^{r*}$ and $\pi_{i-1}^{r*}$.

Leave partitions ($P_{i-1}^1$ and $P_{i}^1$) are used to allow computing fine boundary similarities, whereas boundaries 
from $\pi_{i-2}^{r*}$ and $\pi_{i-1}^{r*}$ are included to enforce previous semantic contours. With this iterative process, 
clusters are robustly propagated through hierarchies in an efficient manner.
\section{Experimental Evaluation}
In this section, we present both qualitative and quantitative evaluations of our multiresolution hierarchical co-clustering (MRHC). As our technique aims to segment sequences with small variations, we use the Video Occlusion/Object Boundary Detection 
Dataset \cite{Stein2009} for evaluation and comparison with state-of-the-art methods in the fields of video segmentation (\cite{Grundmann2010},
\cite{Corso2012}, \cite{Galasso2012}) and co-segmentation (\cite{Joulin2012}, \cite{gunhee2012}). Comparisons have been made using the implementations from respective authors.
In order to asses the contribution of the multi-resolution 
framework (Section \ref{sec:video_co_clustering}), we also evaluate the performance of our algorithm at a single level 
with the best overall results (OURS-SL). 
Moreover,  based on the baseline in \cite{Galasso2013}, we consider a system that propagates labels from regions obtained with \cite{Malik2011} using \cite{Brox2009} (UCM-P).
A random hierarchy created from the leave partitions of \cite{Malik2011} is used as baseline technique.

The dataset includes 30 short sequences (42 objects) with indoor and outdoor scenes, noise and compression artifacts, unconstrained handheld
camera motions and moving objects. For each sequence, the annotation of a single frame is provided as ground truth for segmentation assessment (Section \ref{subsec:segmentation_assessment}). To assess temporal consistency (Section \ref{subsec:temporal_coherence}), we have manually annotated the remaining frames by merging regions from the leave partitions of \cite{Malik2011}. 
The evaluation is performed using two types of measures. First, we use the measures presented in \cite{Galasso2013}: boundary 
precision-recall (BPR) from \cite{Malik2011} and a volume precision-recall metric (VPR). Second,
as in (\cite{Stein2009}, \cite{Glasner2011}), we use \textit{Consistency} as the Jaccard index computed between a set of regions of a partition and the ground-truth and \textit{Efficiency}
as the minimum number of regions requested to obtain a given consistency.


In order to qualitatively assess our technique and to explore its limitations, we also analyze a subset of sequences from the SegTrack v2 Dataset \cite{Fuxin2013}, some of them containing strong deformations and rapid variations.
In all the experiments, hierarchies have been obtained using \cite{Malik2011} and $30$ resolution levels have been created per sequence ranging between
$[40\%,10\%]$ the number of leaf contours ($\beta = 0.1$). 
\subsection{Segmentation assessment}
\label{subsec:segmentation_assessment}
\begin{figure*}[t]
\centering
\includegraphics[width=0.245\textwidth] {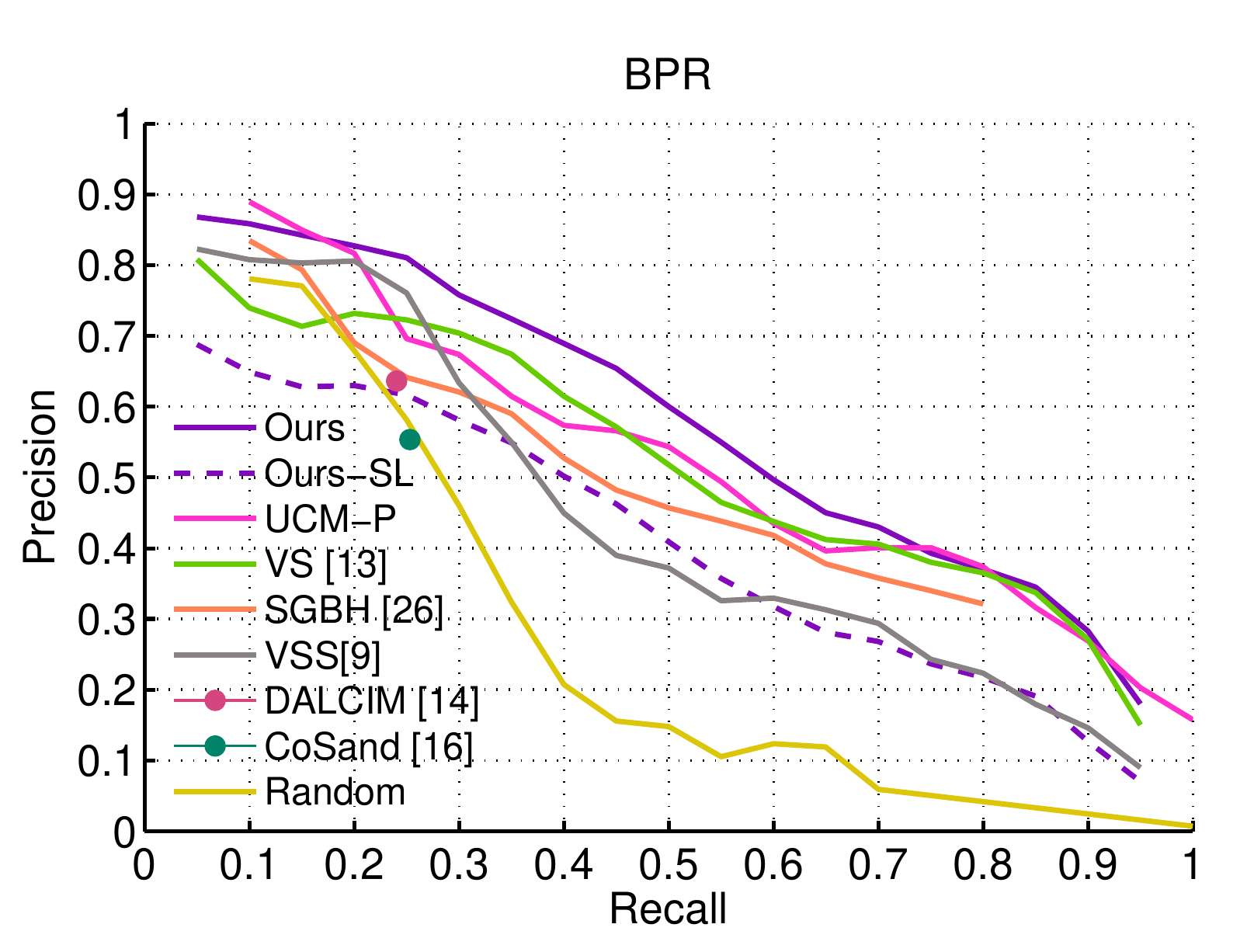}
\includegraphics[width=0.24\textwidth] {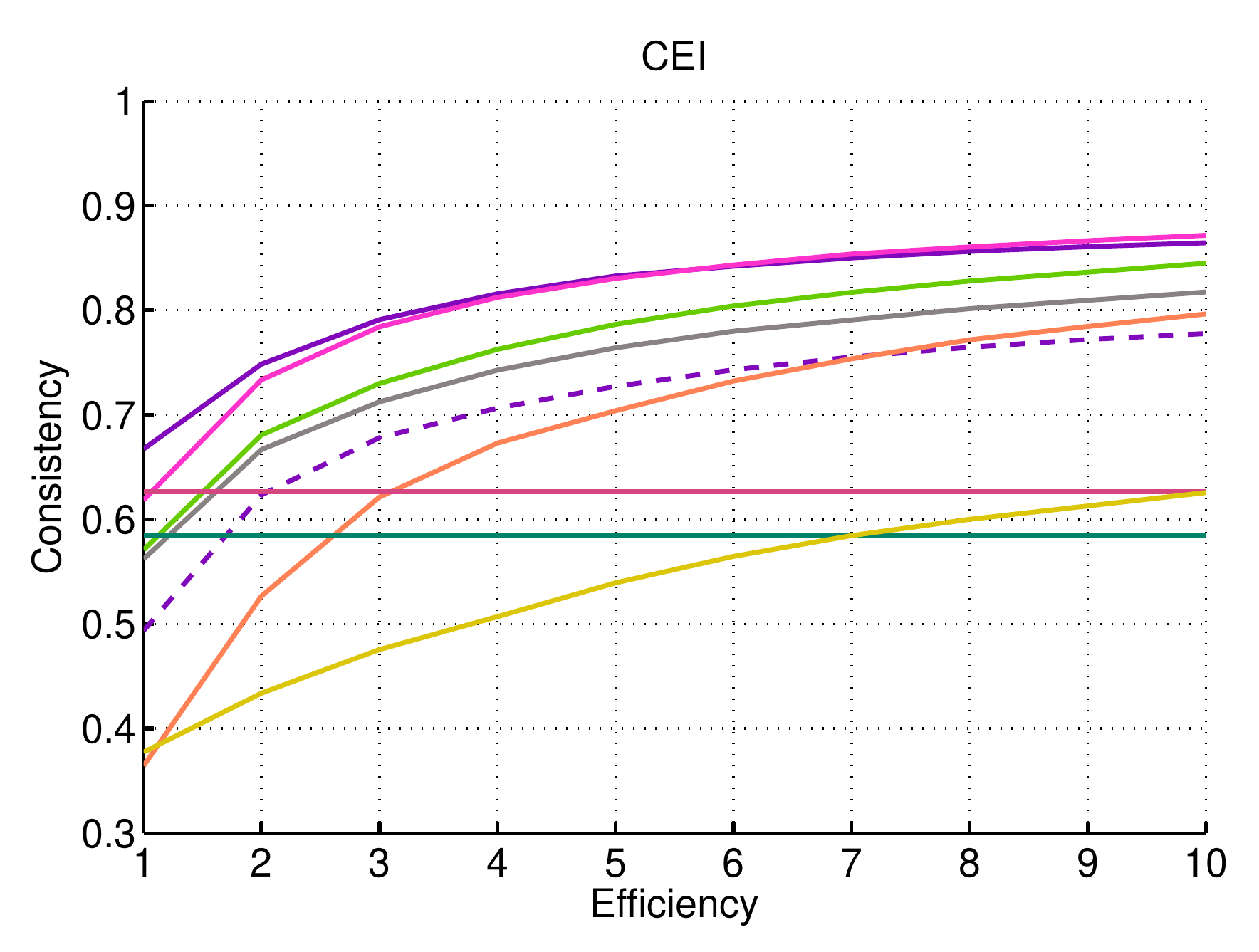}
\includegraphics[width=0.245\textwidth] {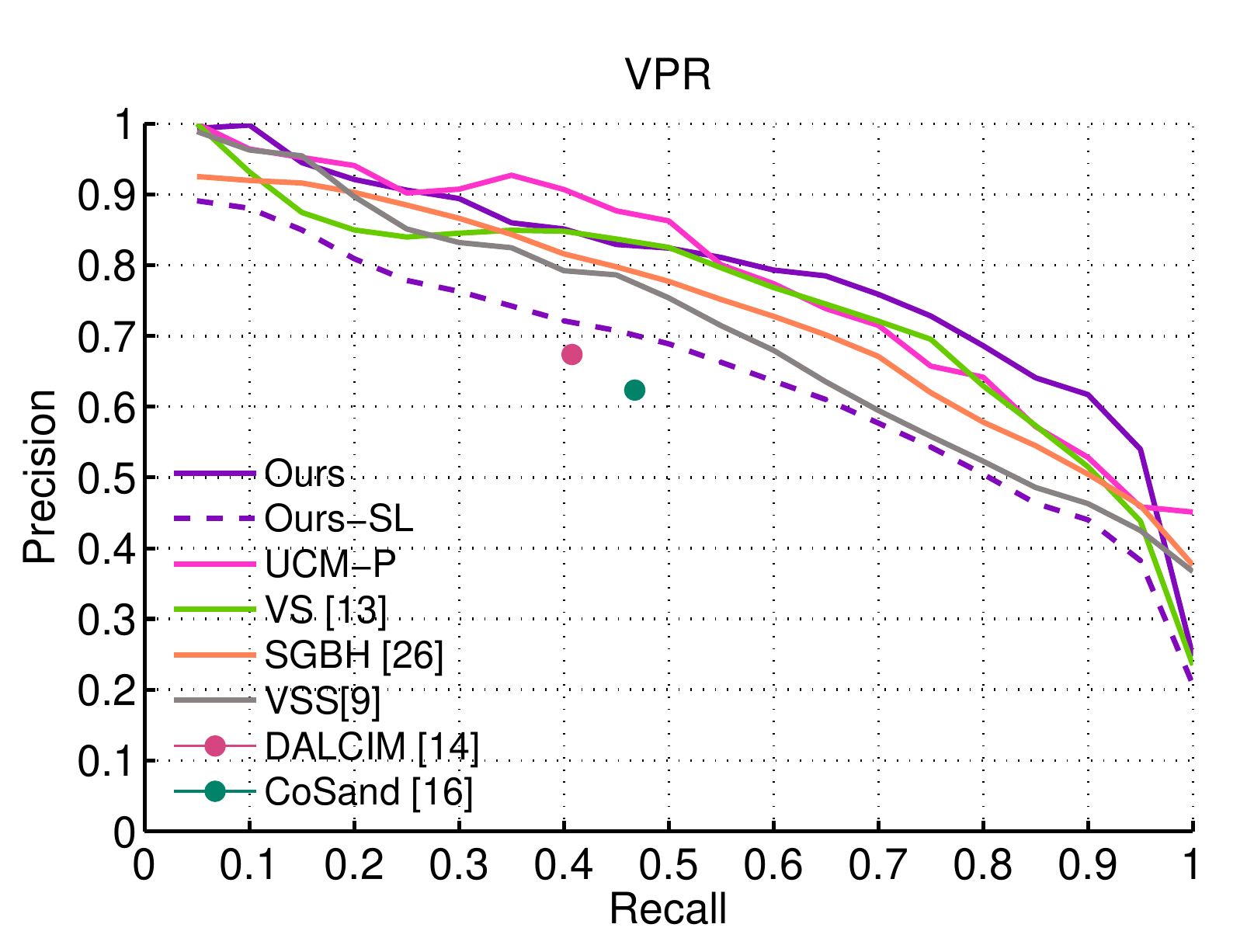}
\includegraphics[width=0.24\textwidth] {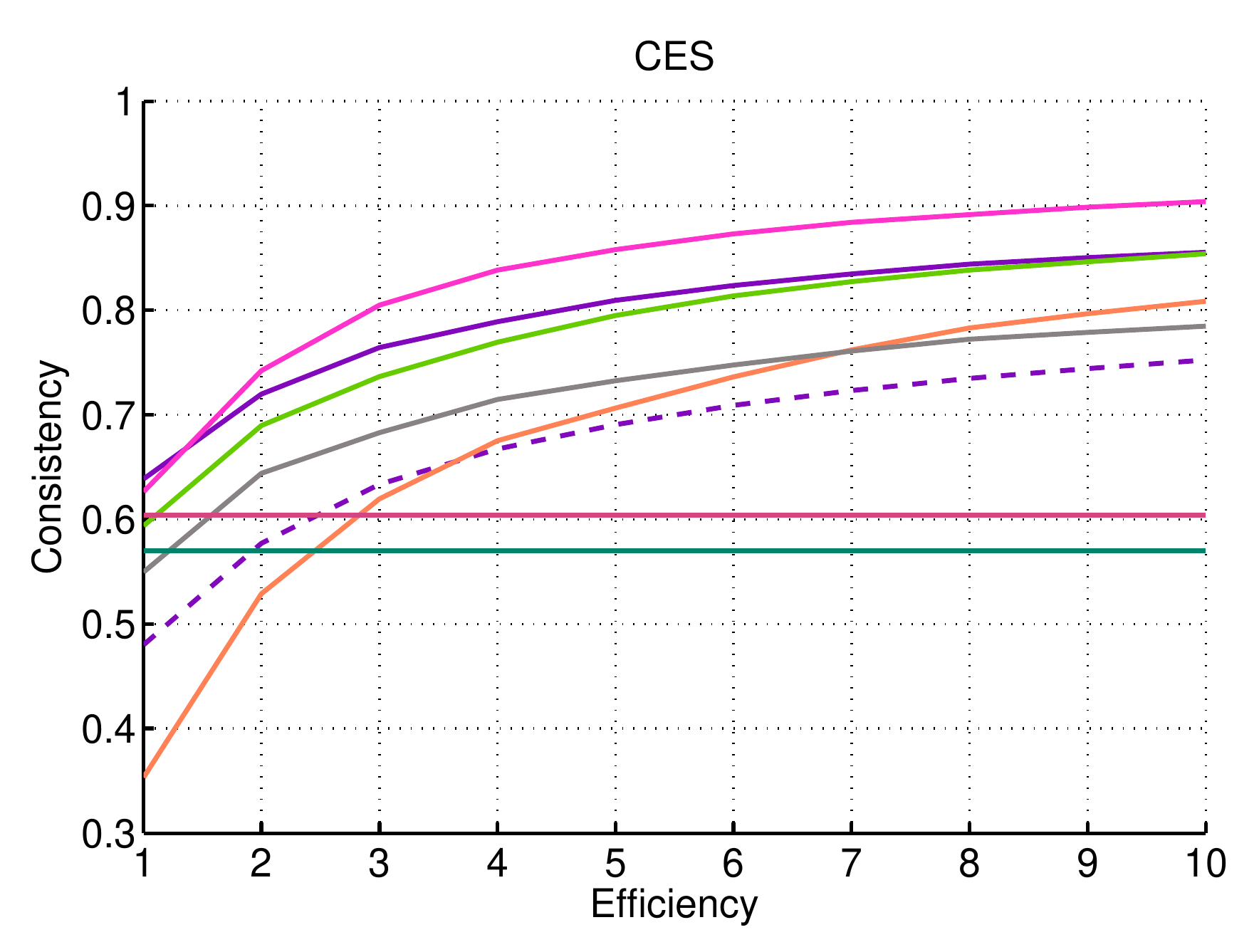}
\caption{Comparison between different methods evaluating their boundary precision-recall (BPR), volume precision-recall (VPR) and 
their consistency for different levels of efficiency both over a single image (CEI) and a sequence (CES).}
\label{fig:total_sequence}
\end{figure*}

In this experiment, we assess the segmentation quality of a given frame. The set of optimal partitions of this frame for all the resolution levels is 
considered. Then, for each efficiency value, the maximum consistency over this set of levels is selected; that is, fixing the number of regions, we 
select through the various resolutions the best Jacard object representation. 
Moreover, the BPR curve is considered to assess the quality of segmentation boundaries (Figure \ref{fig:total_sequence}).
%
%

Co-segmentation results have been obtained fixing the number of clusters with respect to the number of objects in the scene, as proposed by the authors 
(\cite{Joulin2012}, \cite{gunhee2012}). We report the best results for up to a given number of clusters, since consistency does not improve when 
increasing the number of clusters. 
These algorithms 
are competitive when the object is represented with one 
region. Still, our technique obtains better consistency for all efficiency levels.
due to hierarchies and similarities among frames to describe objects. 

Regarding video segmentation algorithms, our technique outperforms the three assessed state-of-the-art methods 
(\cite{Grundmann2010}, \cite{Corso2012}, \cite{Galasso2012}).
In \cite{Corso2012}, colour similarities are used to propagate supervoxels information. In contrast, our description of contours using 
colour, texture and distance measures, obtains better segmentation accuracy and BPR for all precision levels. Although the optical flow 
used in \cite{Grundmann2010} is a powerfull descriptor, it is not enough to accurately segment objects in this type of sequences, specially with a low 
number of regions. As it can be observed in Figure \ref{fig:total_sequence}, in terms of boundaries, their recall is close to our results for large precision 
values. However, in terms of object area, regions selected by our algorithm represent the object with higher accuray. 

Table \ref{tab:comparison} shows the number of objects from the database in which our algorithm obtains better/worse consistency for more than 
$50\%$ of the efficiency levels shown in Figure \ref{fig:total_sequence}. 

\vspace{-0.08cm}
\subsection{Temporal coherence assessment}
\label{subsec:temporal_coherence}
In this section, we extend the previous "efficiency versus consistency" analysis  to the temporal domain, in order to assess the stability of partitions 
along video sequences.

The sequence consistency of a label (temporal cluster) is computed averaging the consistency values obtained at each frame by the region associated to this 
label. Results of the best sequence consistency achieved for all the resolutions, using the number of labels represented by each efficiency level, are 
plotted in Figure \ref{fig:total_sequence}.
In order to complete the analysis, we also present the VPR curve as computed in \cite{Galasso2013}.


As it can be observed, sequence consistency results are very similar to segmentation consistency ones 
(Figure \ref{fig:total_sequence}). This stability shows that all methods correctly maintain the coherence of the partitions along the sequence. These 
results validate the iterative strategies used in \cite{Corso2012} and in our approach (see Section \ref{sec:video_co_clustering}). 
In both volume precision-recall and consistency-efficiency values, our method outperforms the analyzed state-of-the-art approaches and 
only the propagation method based on \cite{Galasso2013} obtains better volume recall for low precision values and better efficiency. This confirms the 
results that were reported in previous works (\cite{Galasso2013}, \cite{Galasso2014}).


A more detailed comparison of the presented algorithms for the objects in the database can be found in Table \ref{tab:comparison}.


\begin{table}
\centering
  \begin{tabular}{|c|cc|cc|cc|}
    \hline
    \multirow{2}{*}{} &
      \multicolumn{2}{c|}{\textbf{Better}} &
      \multicolumn{2}{c|}{\textbf{Worse}} &
      \multicolumn{2}{c|}{\textbf{Inconclusive}} \\
    &  Ref. & Seq. & Ref. & Seq. & Ref. & Seq. \\
    \hline
    \hline
    \cite{Grundmann2010} & 76\% & 64\% & 19\% & 26\% & 5\% & 10\% \\
    \hline
    \cite{Corso2012} & 88\% & 79\% & 7\% & 19\% & 5\% & 2\% \\
    \hline
    \cite{Galasso2012} & 81\% & 74\% & 16\% & 14\% & 3\% & 12\% \\
    \hline
    \cite{Joulin2012} & 88\% & 89\% & 12\% & 9\% & 0\% & 2\% \\
    \hline
    \cite{gunhee2012}& 90\% & 93\% & 10\% & 7\% & 0\% & 0\% \\
    \hline
    OURS-SL & 81\% & 89\% & 12\% & 5\% & 7\% & 6\% \\
    \hline
    UCM-P & 65\% & 34\% & 27\% & 62\% & 8\% & 4\% \\
    \hline
  \end{tabular}
\vspace{0.02cm}
  \caption{Objects of the database for which our algorithm obtains better/worse image (Ref) or sequence (Seq) consistency at more than $50\%$ efficiency levels. 
Otherwise, it is said to be inconclusive.}
\label{tab:comparison}
\end{table}
\subsection{Qualitative assessment}
In this section, we present results on two sequences from the Segtrack v2 database \cite{Fuxin2013} to qualitative evaluate our algorithm. This database allows analyzing the limits of our technique, since video objects in it may undergo strong deformations and rapid movements.

\begin{figure}[t]
\centering
\includegraphics[width=0.23\columnwidth, height=1.55cm] {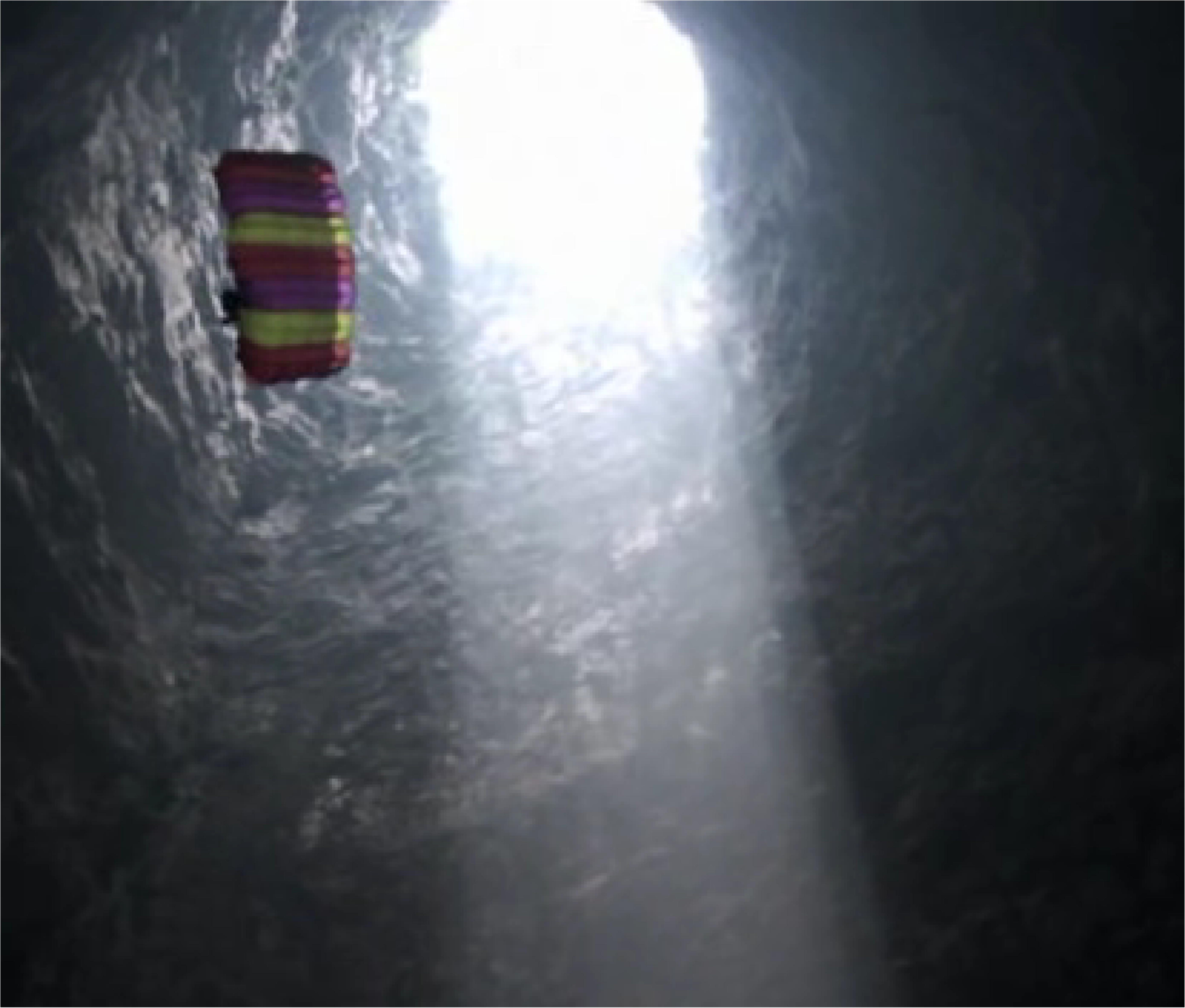}
\includegraphics[width=0.23\columnwidth, height=1.55cm] {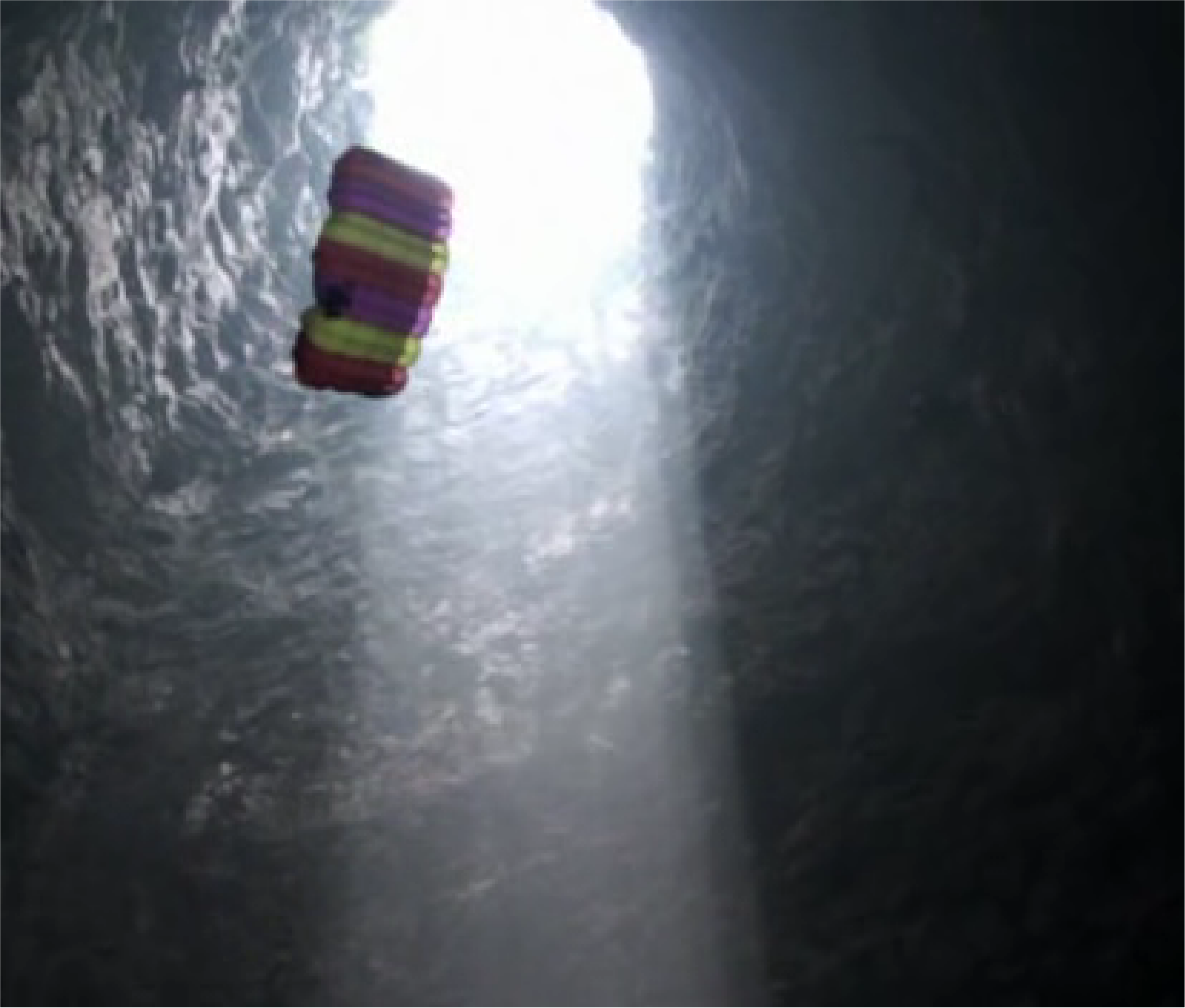}
\includegraphics[width=0.23\columnwidth] {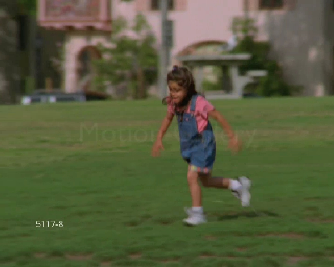}
\includegraphics[width=0.23\columnwidth] {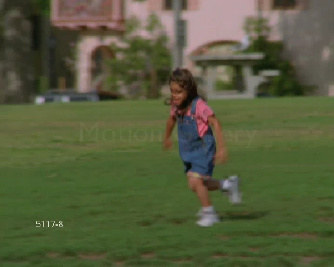}\\

\includegraphics[width=0.23\columnwidth, height=1.55cm] {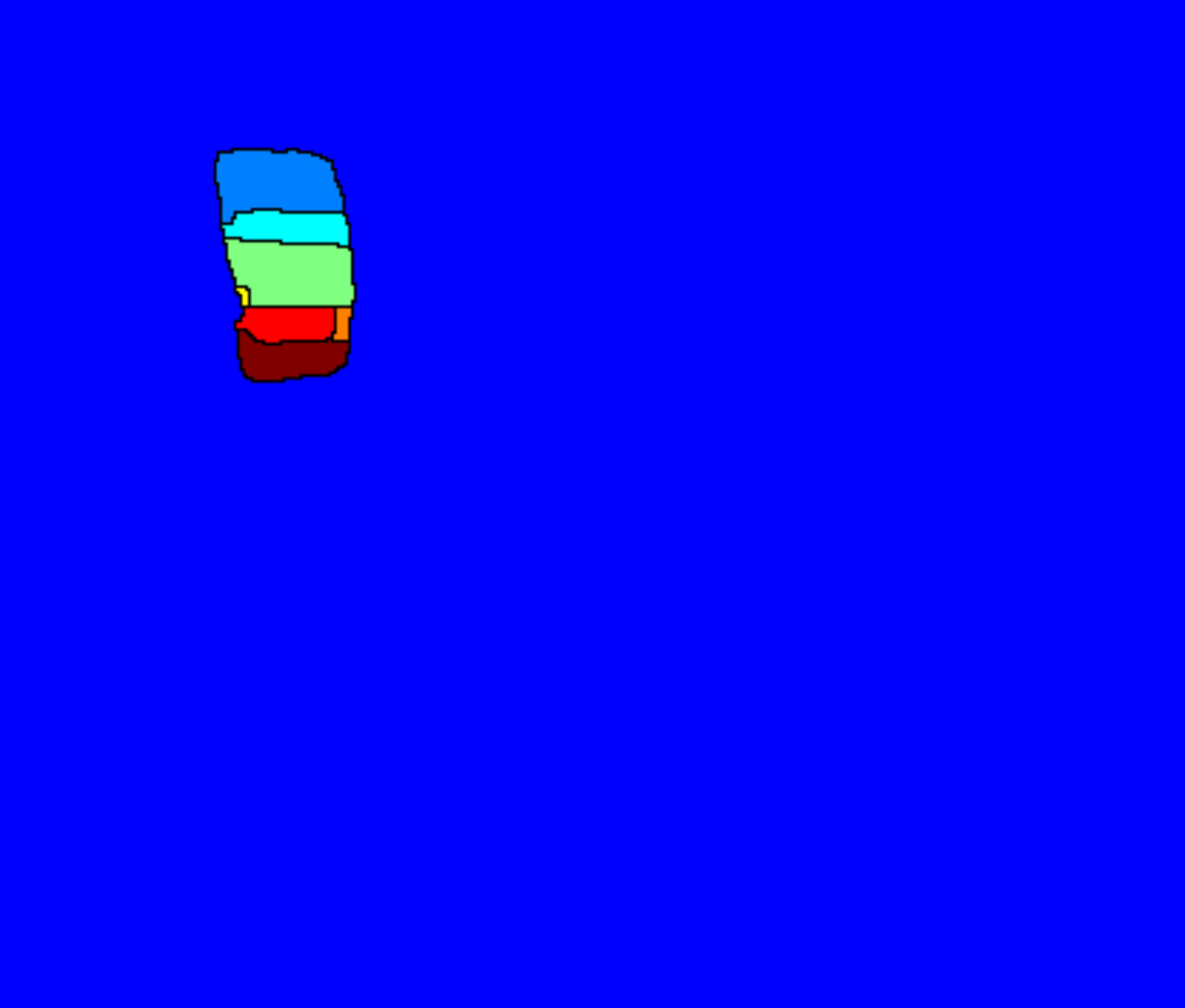}
\includegraphics[width=0.23\columnwidth, height=1.55cm] {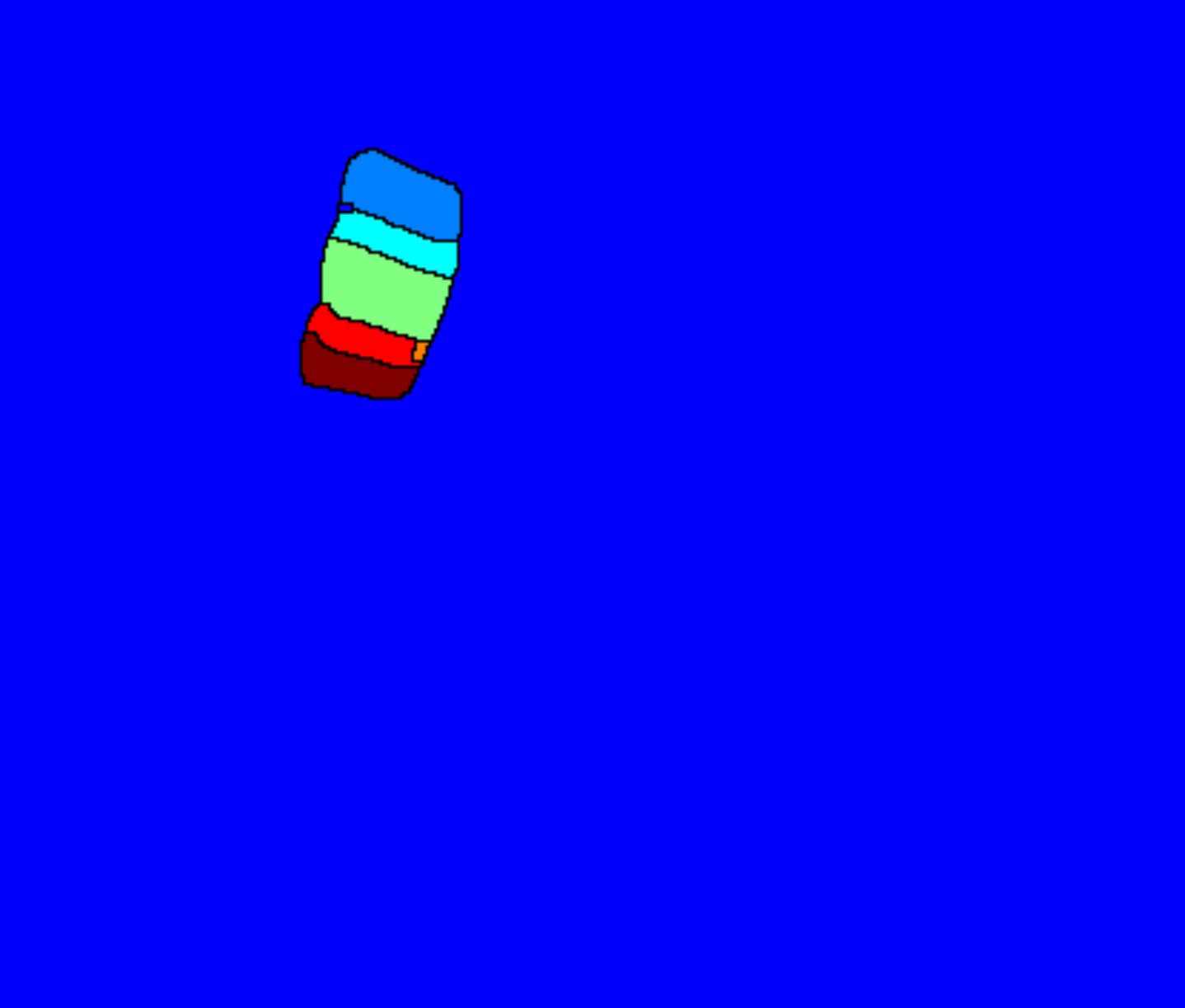}
\includegraphics[width=0.23\columnwidth] {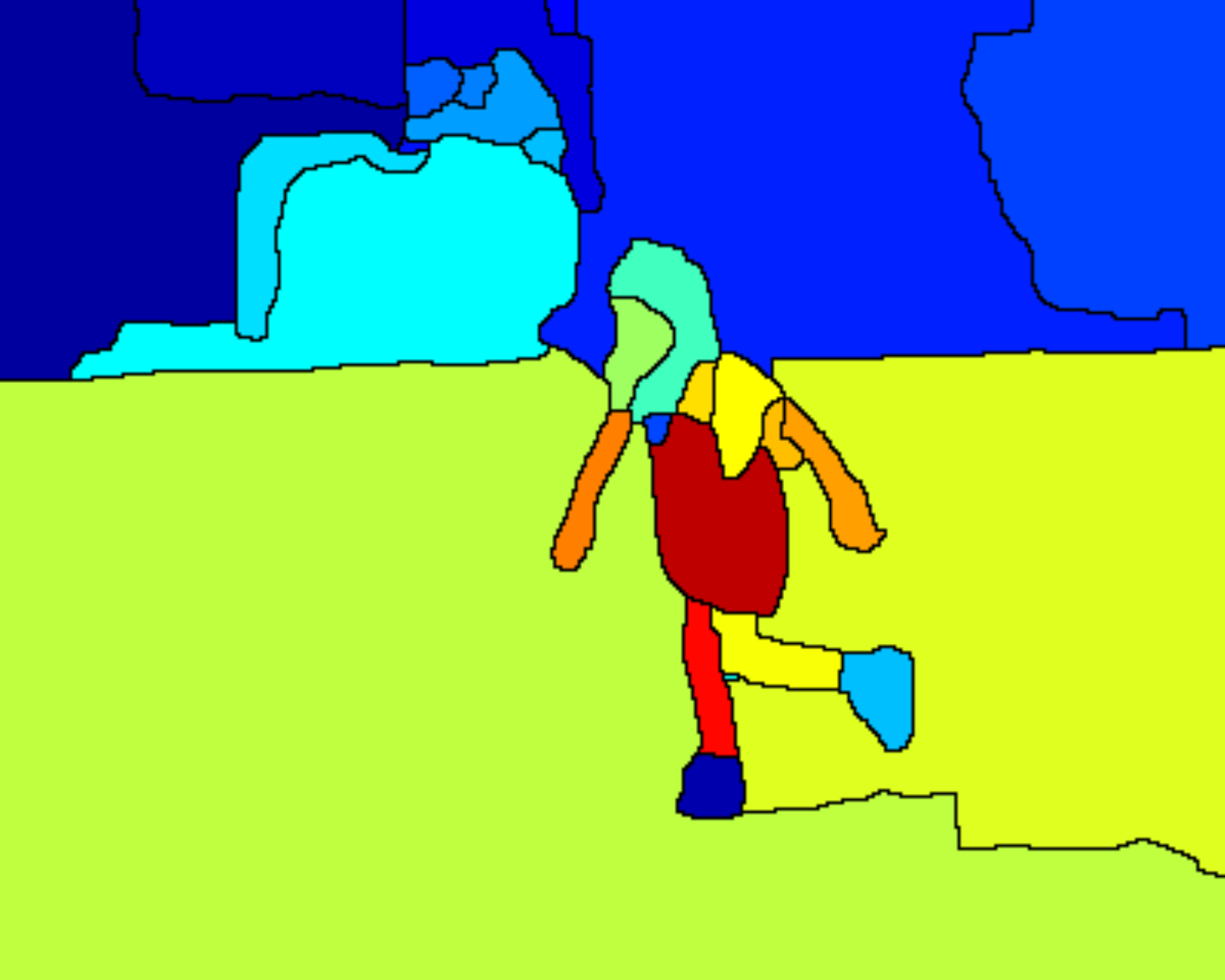}
\includegraphics[width=0.23\columnwidth] {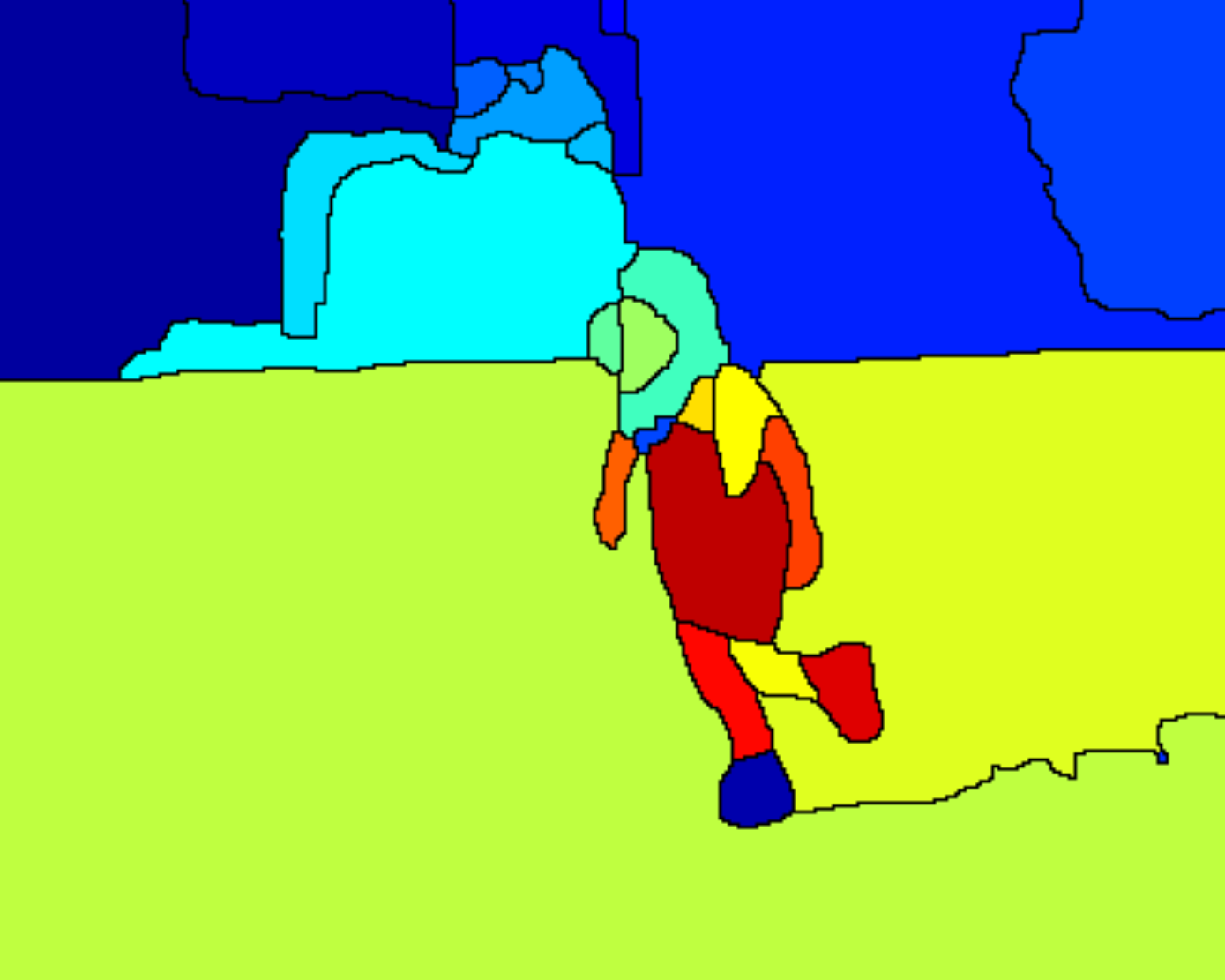}
\caption{Qualitative evaluation of sequences \textit{Parachute} and \textit{Girl} from the SegTrack v2 database. First row: original images. Second
row: result of our iterative video segmentation method.}
\label{fig:qualitative_evaluation}
\end{figure}

Figure \ref{fig:qualitative_evaluation} shows two images of the sequence \textit{Parachute}. 
In it, the parachute is correctly segmented along the sequence at a given resolution. Moreover, its coloured stripes are coherently segmented through 
the sequence. As the object shape gradually changes, our method is able to coherently segment it at several resolution levels along the video.

Figure \ref{fig:qualitative_evaluation} shows two images of the sequence \textit{Girl}. In this sequence, a girl runs and her shape undergoes strong 
deformations due to arm and leg rapid movements. Although the shape of the girl is correctly identified in both partitions as the union of a few regions 
(high consistency at medium efficiency for segmentation), not all its parts have been coherently matched (worse efficiency for temporal coherence).
\vspace{-0.15cm}
\section{Conclusions}
In this work, we have presented a co-clustering framework that creates a coherent region-based multiresolution representation of an image collection, 
by clustering nodes from a collection of independent hierarchies. The co-clustering problem is formulated as a QSAP problem. Inconsistencies commonly 
derived from such optimization problems are avoided modelling the problem through boundary variables and effectively using hierarchical constraints. 
This way, our method robustly creates inter and intra relations between regions from the image collection.

This co-clustering framework has been particularized to obtain a video
segmentation technique that coherently segments scenes with small variations.
We have adopted an iterative strategy that allows reducing the algorithm complexity and memory requirements, while achieving high temporal coherence. 
We have assessed the results over the Video Occlusion/Object Boundary Detection Datase against five SoA techniques and three baseline ones. In all cases, our technique outperforms the SoA methods in video segmentation and co-segmentation 
for this type of sequence in all range of efficiencies. In order to promote reproducible research, all the resources of this project (code, results and 
evaluation protocols) are publicly available.

{\small
\bibliographystyle{ieee}
\bibliography{biblio}

\begin{thebibliography}{10}\itemsep=-1pt

\bibitem{Malik2011}
P.~Arbelaez, M.~Maire, C.~Fowlkes, and J.~Malik.
\newblock Contour detection and hierarchical image segmentation.
\newblock {\em IEEE Trans. Pattern Anal. Mach. Intell.}, 33(5):898--916, May
  2011.

\bibitem{Arbelaez2014}
P.~Arbelaez, J.~Pont-Tuset, J.~Barron, F.~Marques, and J.~Malik.
\newblock Multiscale combinatorial grouping.
\newblock In {\em Computer Vision and Pattern Recognition (CVPR)}, 2014.

\bibitem{Bansal04}
N.~Bansal, A.~Blum, and S.~Chawla.
\newblock Correlation clustering.
\newblock {\em Machine Learning}, 56(1-3):89--113, 2004.

\bibitem{Bhattacharyya1943}
A.~Bhattacharyya.
\newblock {On a measure of divergence between two statistical populations
  defined by their probability distributions}.
\newblock {\em Bulletin of Calcutta Mathematical Society}, 1943.

\bibitem{Brendel2009}
W.~Brendel and S.~Todorovic.
\newblock Video object segmentation by tracking regions.
\newblock In {\em Computer Vision, 2009 IEEE 12th International Conference on},
  pages 833--840, Sept 2009.

\bibitem{Brox2009}
T.~Brox, C.~Bregler, and J.~Malik.
\newblock Large displacement optical flow.
\newblock In {\em Computer Vision and Pattern Recognition, 2009. CVPR 2009.
  IEEE Conference on}, pages 41--48, June 2009.

\bibitem{Charikar2005}
M.~Charikar, V.~Guruswami, and A.~Wirth.
\newblock Clustering with qualitative information.
\newblock In {\em Foundations of Computer Science, 2003.}, pages 524--533.

\bibitem{Wei2013}
W.-C. Chiu and M.~Fritz.
\newblock Multi-class video co-segmentation with a generative multi-video
  model.
\newblock In {\em Computer Vision and Pattern Recognition (CVPR), 2013 IEEE
  Conference on}, pages 321--328, June 2013.

\bibitem{Galasso2012}
F.~Galasso, R.~Cipolla, and B.~Schiele.
\newblock Video segmentation with superpixels.
\newblock In {\em Asian Conference on Computer Vision}, 2012.

\bibitem{Galasso2014}
F.~Galasso, M.~Keuper, T.~Brox, and B.~Schiele.
\newblock Spectral graph reduction for efficient image and streaming video
  segmentation.
\newblock In {\em IEEE Conf. on Computer Vision and Pattern Recognition}, 2014.

\bibitem{Galasso2013}
F.~Galasso, N.~S. Nagaraja, T.~J. Cardenas, T.~Brox, and B.~Schiele.
\newblock A unified video segmentation benchmark: Annotation, metrics and
  analysis.
\newblock In {\em IEEE International Conference on Computer Vision}, 2013.

\bibitem{Glasner2011}
D.~Glasner, S.~Vitaladevuni, and R.~Basri.
\newblock Contour-based joint clustering of multiple segmentations.
\newblock In {\em CVPR}, 2011.

\bibitem{Grundmann2010}
M.~Grundmann, V.~Kwatra, M.~Han, and I.~Essa.
\newblock Efficient hierarchical graph based video segmentation.
\newblock {\em IEEE CVPR}, 2010.

\bibitem{Joulin2012}
A.~Joulin, F.~Bach, and J.~Ponce.
\newblock Multi-class cosegmentation.
\newblock In {\em Computer Vision and Pattern Recognition (CVPR), 2012 IEEE
  Conference on}, pages 542--549, June 2012.

\bibitem{Kim2012}
E.~Kim, H.~Li, and X.~Huang.
\newblock A hierarchical image clustering cosegmentation framework.
\newblock In {\em Computer Vision and Pattern Recognition (CVPR), 2012 IEEE
  Conference on}, pages 686--693, June 2012.

\bibitem{gunhee2012}
G.~Kim and E.~P. Xing.
\newblock On multiple foreground cosegmentation.
\newblock In {\em 25th IEEE Conference on Computer Vision and Pattern
  Recognition (CVPR 2012)}, 2012.

\bibitem{Fuxin2013}
F.~Li, T.~Kim, A.~Humayun, D.~Tsai, and J.~M. Rehg.
\newblock Video segmentation by tracking many figure-ground segments.
\newblock In {\em ICCV}, 2013.

\bibitem{Maire2008}
M.~Maire, P.~Arbelaez, C.~Fowlkes, and J.~Malik.
\newblock Using contours to detect and localize junctions in natural images.
\newblock In {\em Computer Vision and Pattern Recognition, 2008. CVPR 2008.
  IEEE Conference on}, pages 1--8, June 2008.

\bibitem{Paris2008}
S.~Paris.
\newblock Edge-preserving smoothing and mean-shift segmentation of video
  streams.
\newblock In D.~Forsyth, P.~Torr, and A.~Zisserman, editors, {\em Computer
  Vision – ECCV 2008}, volume 5303 of {\em Lecture Notes in Computer
  Science}, pages 460--473. Springer Berlin Heidelberg, 2008.

\bibitem{Rother2006}
C.~Rother, T.~Minka, A.~Blake, and V.~Kolmogorov.
\newblock Cosegmentation of image pairs by histogram matching - incorporating a
  global constraint into mrfs.
\newblock In {\em Computer Vision and Pattern Recognition, 2006 IEEE Computer
  Society Conference on}, volume~1, pages 993--1000, June 2006.

\bibitem{Rubio2012}
J.~Rubio, J.~Serrat, and A.~López.
\newblock Video co-segmentation.
\newblock In K.~Lee, Y.~Matsushita, J.~Rehg, and Z.~Hu, editors, {\em Computer
  Vision – ACCV 2012}, volume 7725 of {\em Lecture Notes in Computer
  Science}, pages 13--24. Springer Berlin Heidelberg, 2013.

\bibitem{Salembier00}
P.~Salembier and L.~Garrido.
\newblock Binary partition tree as an efficient representation for image
  processing, segmentation and information retrieval.
\newblock {\em IEEE transactions on image processing},
  9(4):561{\textendash}576, 2000.

\bibitem{Stein2009}
A.~Stein and M.~Hebert.
\newblock Occlusion boundaries from motion: Low-level detection and mid-level
  reasoning.
\newblock {\em International Journal on Computer Vision}, 82(2):325--357, April
  2009.

\bibitem{Vitaladevuni2010}
S.~Vitaladevuni and R.~Basri.
\newblock Co-clustering of image segments using convex optimization applied to
  em neuronal reconstruction.
\newblock In {\em CVPR}, 2010.

\bibitem{Corso2013}
C.~Xu, S.~Whitt, and J.~J. Corso.
\newblock Flattening supervoxel hierarchies by the uniform entropy slice.
\newblock In {\em Proceedings of the IEEE International Conference on Computer
  Vision}, 2013.

\bibitem{Corso2012}
C.~Xu, C.~Xiong, and J.~J. Corso.
\newblock Streaming hierarchical video segmentation.
\newblock In {\em Proceedings of European Conference on Computer Vision}, 2012.

\end{thebibliography}
}

\end{document}